\pdfoutput=1
\documentclass[10pt, logo, onecolumn, copyright]{nv}
\usepackage{graphicx}
\definecolor{nvidiagreen}{HTML}{76B900}

\usepackage{amsmath,amsfonts,bm}

\def\eqref#1{equation~\ref{#1}}

\def\1{\bm{1}}

\DeclareMathAlphabet{\mathsfit}{\encodingdefault}{\sfdefault}{m}{sl}
\SetMathAlphabet{\mathsfit}{bold}{\encodingdefault}{\sfdefault}{bx}{n}

\usepackage[nameinlink]{cleveref}
\usepackage{hyperref}
\usepackage[square,sort,comma,numbers]{natbib}
\usepackage{url}
\usepackage{xspace}
\usepackage{lipsum}
\usepackage{multirow}
\usepackage{multicol}
\usepackage{colortbl}
\usepackage{listings}
\usepackage{enumitem}
\usepackage{tabularx}
\usepackage{booktabs}
\usepackage{array}
\usepackage{caption}
\usepackage{enumitem}
\usepackage{caption}
\usepackage{circledsteps}
\usepackage{vcell}
\usepackage{bbm}
\usepackage{wrapfig}
\usepackage[many]{tcolorbox}
\usepackage{booktabs}
\usepackage{ulem}
\usepackage{makecell}

\crefformat{section}{\S#2#1#3} 
\crefformat{subsection}{\S#2#1#3}
\crefformat{subsubsection}{\S#2#1#3}

\newcommand{\reffig}[1]{Figure~\ref{#1}}
\newcommand{\refsec}[1]{\S\ref{#1}} 
\newcommand{\reftab}[1]{Table~\ref{#1}}

\def\ie{\textit{i.e.}\xspace}

\newcommand{\RNum}[1]{\uppercase\expandafter{\romannumeral #1\relax}}

\definecolor{myyellow}{RGB}{249, 231, 173}
\definecolor{mygreen}{RGB}{233, 242, 230}
\definecolor{mygrey}{RGB}{250, 250, 250}
\definecolor{myblue}{RGB}{245, 248, 250}
\definecolor{communication}{RGB}{224, 242, 254}
\definecolor{situation}{RGB}{220, 252, 231}
\definecolor{planning}{RGB}{255, 237, 213}
\definecolor{environment}{RGB}{243, 232, 255}
\definecolor{personalization}{RGB}{254, 226, 226}

\definecolor{codegreen}{rgb}{0,0.6,0}
\definecolor{codegray}{rgb}{0.5,0.5,0.5}

\definecolor{backcolour}{RGB}{245,248,250}
\definecolor{emph}{RGB}{166,88,53}
\definecolor{nightblue}{RGB}{9,49,105}
\definecolor{keywords}{RGB}{207,33,46}
\definecolor{lightpurple}{RGB}{130,81,223}

\lstdefinestyle{mystyle}{
    backgroundcolor=\color{backcolour},   
    commentstyle=\color{codegreen},
    keywordstyle=\color{keywords},
    stringstyle=\color{nightblue},
    basicstyle=\fontsize{7}{8}\ttfamily,
    breakatwhitespace=true,         
    breaklines=true,                 
    captionpos=b,                    
    keepspaces=true,                 
    numberstyle=\tiny\color{codegray},
    numbersep=2pt,                  
    showspaces=false,                
    showstringspaces=false,
    showtabs=false,                  
    tabsize=2,
    emph={dspy},
    emphstyle={\color{lightpurple}},
    linewidth=1\columnwidth,
    frame=tb,    
    xrightmargin=0pt,
    xleftmargin=0.23cm,
    numbers=left,
    aboveskip=0.2cm,
    belowskip=0.1cm,
}

\newtcolorbox{promptbox}[1]{
    enhanced,
    breakable,
    boxrule=1pt,  
    fontupper=\small,
    fonttitle=\bfseries\color{white},
    arc=3pt,  
    rounded corners,
    colframe=black,
    colbacktitle=gray!50!black,
    colback=mygrey,
    title=#1,
    left=2mm,  
    right=2mm,  
    top=1mm,  
    bottom=1mm  
}

\newtcolorbox{agentbox}[1]{
    enhanced,
    breakable,
    boxrule=1pt,  
    fontupper=\small,
    fonttitle=\bfseries\color{white},
    arc=3pt,  
    rounded corners,
    colframe=black,
    colbacktitle=gray!50!black,
    colback=myblue,
    title=#1,
    left=2mm,  
    right=2mm,  
    top=1mm,  
    bottom=1mm  
}

\usepackage[toc,page,header]{appendix}
\usepackage{minitoc}
\title{Deep Personalization: Continuous Context Observation for Local Proactive Agents}

\correspondingauthor={%
  \footnotesize\textnormal{%
  *Work done during internship at NVIDIA.\\
  $^\dagger$Work done at NVIDIA.\\%
  }%
}

\author{
\parbox{\linewidth}{
{\Authfont
Yijia Shao\textsuperscript{1,2*}, Song Bian\textsuperscript{1},
Junxian Guo\textsuperscript{1,3*}, Jindong Jiang\textsuperscript{1},
Zhen Dong\textsuperscript{1},
Huizi Mao\textsuperscript{4$\dagger$},\par}
{\Authfont
Diyi Yang\textsuperscript{2}, Song Han\textsuperscript{1,3}\par}

\vspace{1em}

{\Authfont
\textsuperscript{1}NVIDIA \qquad
\textsuperscript{2}Stanford University \qquad
\textsuperscript{3}Massachusetts Institute of Technology \qquad
\textsuperscript{4}OpenAI\par}
}
}

\begin{abstract}
\vspace{-10pt}
\section*{Abstract}

As AI agents become increasingly capable, users must still recognize when and how to ask for help, leaving many opportunities for useful support unrealized. Proactive agents hold the promise to close this gap, but effective proactivity requires understanding users’ real-time activities and personalizing intervention timing so that assistance is offered when it is likely to be helpful rather than disruptive. We introduce \textbf{deep personalization}, a closed-loop process that connects continuous context observation with continual agent adaptation: an agent observes user activity, selectively retains long-term context, and adapts intervention decisions from implicit feedback. We initiate deep personalization in \textbf{Coco}, a proactive desktop agent, by having a separate Observer Agent that monitors desktop activity, maintains compact long-term user modeling context, and learns when assistance is desirable from implicit signals such as engagement, dismissal, explicit requests, and repeated work patterns. We further develop efficient on-device strategies to enable the Observer Agent to run on local AI compute platforms without streaming raw activity data off-device. For evaluation at scale without collecting sensitive real-user activity, we introduce a data-synthesis pipeline that transforms existing computer-use trajectories into \textbf{OSWorld-Persona}, a benchmark with persona-conditioned intervention labels. On OSWorld-Persona, deep personalization improves macro F1 across three model backbones from 0.12-0.19 under direct prompting to 0.35-0.48 after adaptation, a 2.4-3.0$\times$ relative gain. In a week-long user study, we observe that Coco improves its macro-averaged F1 score from 0.199 to 0.515. These results suggest that on-device deep personalization can make proactive agents more timely and increasingly aligned with each user’s everyday work.

\end{abstract}

\begin{document}

\maketitle

\begin{figure*}[h]
  \centering
  \includegraphics[width=1\textwidth]{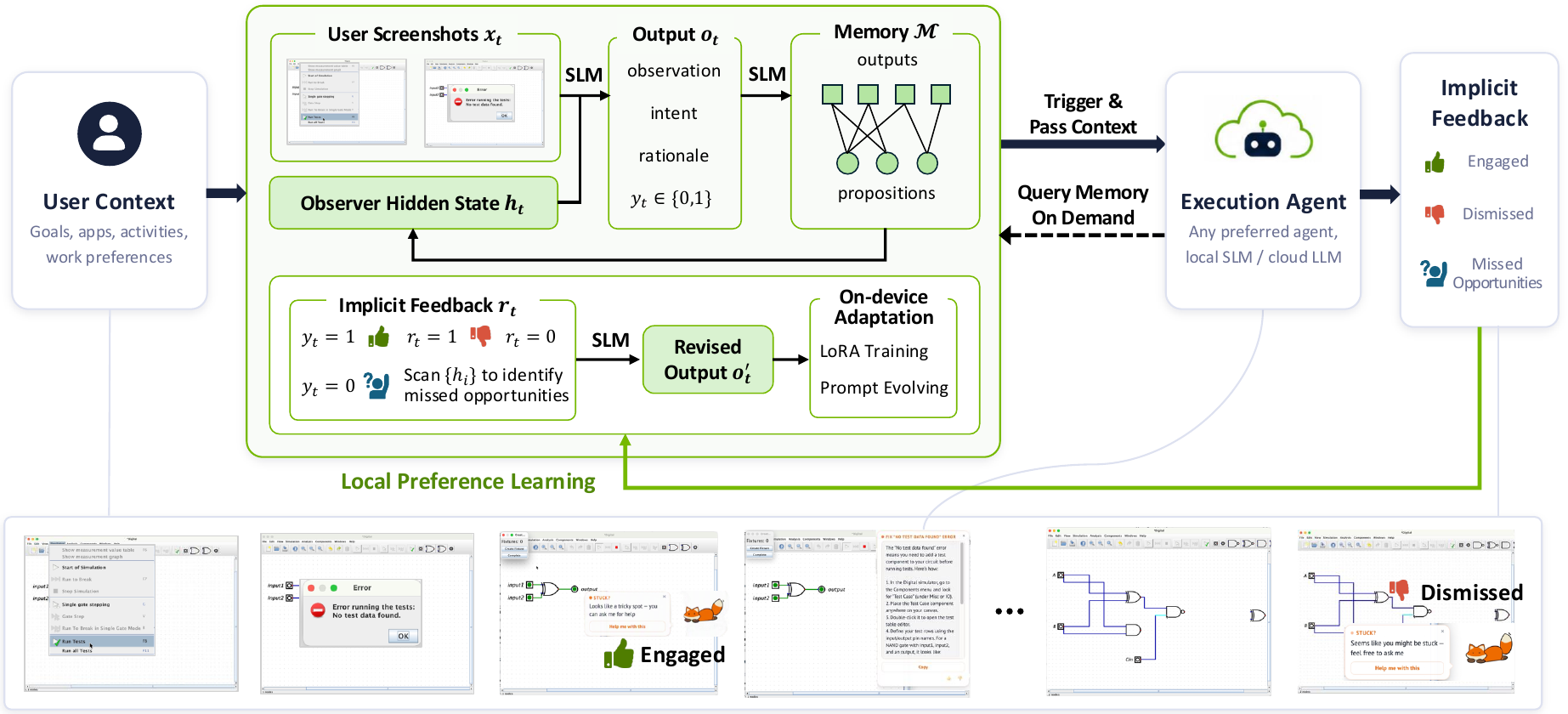}
\caption{\textbf{Overview of Coco, a personalized proactive assistant.} Coco realizes deep personalization by closing the loop of continuous context observation, long-term user modeling and self-adaptation from implicit user feedback with an Observer Agent.
Demo of the Observer Agent running Qwen3.5-35B-A3B-FP8 on NVIDIA DGX Spark: \url{https://youtu.be/fe7MSjoJqZM}
}
\vspace{-1em}
\label{fig:overview}
\end{figure*}


\section{Introduction}
Large language models (LLMs) and LLM-based agents are becoming increasingly capable, transforming how people work~\citep{shao2024assisting,liu2025largelanguagemodelbasedagents,sager2026comprehensive}. Yet most agents today remain reactive: users must first recognize when AI could help, then initiate interaction through explicit instructions. This potentially limits the benefits of increasingly capable agents to those who already know when and how to engage them~\citep{zhang2026college,angrisani2026gaps}. Proactive agents offer a different paradigm, in which agents recognize opportunities for support and initiate interaction itself. This vision has a long history in intelligent interfaces~\citep{maes1994agents,2074094.2074124} and is recently re-emerging with modern LLM-based agents~\citep{lu2025proactive,yang2026fingertip,hu2026anticipate,shaikh2025creating}. However, proactivity requires an agent to decide not only what it can do, but whether its help is wanted at a particular moment. An intervention that is useful for one user may be redundant or disruptive for another, and even the same user may prefer different levels of assistance across tasks and situations~\citep{horvitz1999principles}.

We argue that addressing this problem requires moving beyond personalization based on static profiles or explicit configuration. What proactivity requires is \textbf{deep personalization}: a close-loop process in which an agent continuously observes user activities, selectively retains user context, makes personalized intervention decisions, and adapts itself from implicit user feedback. Such an agent builds its understanding of the user from everyday interaction---what they are working on, how they work, what assistance they value, and when they prefer to be left alone---and keeps revising it as it observes new activity and how its interventions land.

We present \textbf{Coco}, a personalized proactive agent that realizes deep personalization through three specific technical contributions. First, at its core is an Observer Agent that continuously processes desktop activities, maintains a compact representation of the user's ongoing work, and predicts whether proactive assistance would be valuable (\refsec{sec:observer}). Second, a long-term memory module that accumulates context across observations, maintaining a hidden state that is updated with each new observation and used to condition downstream assistance (\refsec{sec:memory}). Third, a closed-loop adaptation mechanism that treats user behavior such as acceptance, rejection, explicit requests, and repetitive-work patterns as implicit supervision signals to continually refine the Observer Agent via evolutionary prompt optimization and LoRA fine-tuning (\refsec{sec:training}). Together, these components enable Coco to learn each user's preferred intervention timing without manual annotation. As there is an inherent tension between personalization and data exposure, we develop an efficient deployment strategy combining change-aware frame selection with observation-first scheduling (\refsec{sec:deployment}) to keep deep personalization on device. We demonstrate that the Observer Agent can run entirely on a single consumer device, such as an RTX 5090 or DGX Spark, with latency suitable for processing user activity over multi-second windows.


To evaluate personalized proactive agents, we introduce \textbf{OSWorld-Persona}, a new benchmark for deep personalization that we construct via a novel data synthesis pipeline. Rather than collecting real user data which raises privacy concerns and limits reproducibility, we leverage existing computer-use benchmarks and use VLM-based user simulators conditioned on distinct personas to generate personalized labels for intervention timing appropriateness. Experiments on OSWorld-Persona show that personalized adaptation is essential for proactive support: direct prompting rarely identifies useful intervention opportunities (macro recall 0.08-0.19 and F1 below 0.19), while personalization via evolutionary prompt optimization or LoRA training raises macro F1 by 2.4-3.0$\times$ across base models and reaches up to 0.77 F1 on one of the personas. We further evaluate Coco on real user workloads in a week-long user study with 10 participant. Coco improves its macro-averaged F1 score from 0.199 to 0.515, demonstrating the effectiveness of deep personalization in the wild.

In summary, our contributions include:
\begin{itemize}
\vspace{-0.5em}
    \item {We introduce \textbf{Coco}, a personalized proactive agent built with deep personalization. 
    Its Observer Agent continuously observes the user's digital context, learns when assistance is desirable from implicit feedback, and provides relevant context to the Execution Agent.}
    \item {We develop an efficient on-device deployment strategy that runs the Observer Agent locally for both time-sensitive screen understanding and personalization workloads, demonstrating that local compute can support privacy-preserving personalization in a large agentic system.}
    \item {We introduce a data curation pipeline for deriving synthetic personalization benchmarks from computer-use datasets and use it to curate OSWorld-Persona.\footnote{Deep Personalization code and OSWorld-Persona data are available at \url{https://github.com/NVlabs/deep-personalization}.} We further instantiate Coco as a desktop app and validate its effectiveness in a week-long user study.\footnote{Desktop proactive agent integration is available at \url{https://github.com/collaborative-agents/coco}.
    }}
\vspace{-0.5em}
\end{itemize}

\section{Learning Personalized Agents from Continuous Observation}
\label{sec:background}
We study the problem of learning a personalized AI agent for an individual user. Personalized behavior depends not only on relatively stable attributes such as stated preferences, but also on how a user works, what assistance they value, and when they prefer not to be interrupted. This requires learning a personalized agent from the user's own behavior. 
While various formulations and experimental setups have been proposed for personalized AI agents~\citep{yang2026fingertip,liu2026sensing,lyu2026personalalign,liang2026learning}, the core problem can be stated as follows: given an AI agent with minimal prior knowledge, enable it to learn appropriate behavior from the user and their interactions~\citep{maes1994agents,xu2026toward}. Formally, let $s_t$ denote the environment state at timestep $t$ and $h_t$ denote the agent's internal state. The agent's behavior is abstracted as a policy $\pi(a_t \mid s_t, h_t)$ that selects action $a_t$ based on the environment state and internal state. Given feedback $r_t = \mathcal{R}(a_t)$ obtained from user $u$, personalization aims to find a learning algorithm $\mathcal{L}$ that yields the optimal user-conditioned policy from the interaction history:
\begin{equation}
    \pi_u^* = \mathcal{L}(\tau^{(u)}) \propto \mathbb{E}_{\tau^{(u)}\sim\pi}\left[\Sigma_{t=1}^{T}\gamma^{t-1}r_t\right],
\end{equation}
where $\tau^{(u)}=(s_1,h_1,a_1,...,s_T,h_T,a_T)$ is the trajectory induced by $\pi$ for user $u$, and $\gamma\in(0,1]$ is a discount factor.

A central challenge in this formulation is obtaining informative $s_t$ and $r_t$ without repeatedly asking the user to explicitly specify their context or provide supervision. We consider \textbf{continuous context observation} as a natural source of both. For example, in digital environments, the agent can observe a representation of what the user themselves sees and does (\ie, desktop screenshots and interaction history), and use this stream to infer ongoing activities and maintain $h_t$. The same interaction stream can provide feedback after the agent acts: editing an AI-generated draft, engaging with a proactive suggestion, dismissing it, or subsequently requesting assistance can all provide evidence about whether the agent's behavior matched the user's needs. Continuous observation therefore creates a \textbf{\textit{closed-loop} deep personalization process} in which user activity provides context for an agent decision, and subsequent activity provides supervision for improving future decisions.



\section{Method}
We present \textbf{Coco}, a proactive assistant built on \uline{co}ntinuous \uline{co}ntext observation to personalize intervention timing. 
At its core, an Observer Agent transforms high-frequency desktop observations into compact structured representations and predict whether proactive assistance would be valuable (\refsec{sec:observer}). Coco achieves deep personalization through long-term user modeling (\refsec{sec:memory}) and continual observer adaptation from implicit feedback (\refsec{sec:training}). \reffig{fig:overview} illustrates the system overview.

\subsection{Observer Agent}
\label{sec:observer}
Deep personalization requires the agentic system to maintain a coherent, evolving understanding of the user across time. Central to our personalization framework is the Observer Agent $\mathcal{O}$, which serves two tightly coupled roles: continuously distilling high-frequency desktop observations into compact structured representations of user activity, and deciding when proactive assistance would be valuable. Following the formulation in~\refsec{sec:background}, let $s_t$ denote the user-observable environment state, \ie, desktop screenshots together with their timestamps, and $h_{t-1}$ denote the personalized context accumulated before timestep $t$. At timestep $t$, the Observer Agent processes $(s_t,h_{t-1})$ and produces a structured observation $o_t$ and an intervention decision $y_t$:
\begin{equation}
    (y_t, h_t) = \mathcal{O}(s_t, h_{t-1}),
\label{eq:observer_agent}
\end{equation}
where $y_t\in\{0,1\}$ denotes whether proactive assistance should be offered. When $y_t=1$, Coco invokes the downstream Execution Agent; when $y_t=0$, it continues passive observation.

We parameterize $\mathcal{O}$ as a multimodal language model capable of processing at least text and image inputs. 
Given a single prompt, $\mathcal{O}$ produces an output $o_t$ consisting of four structured output fields: 
(1)~\textit{observation}, a compact incremental update distilling valuable information  from the user's recent computer-use activity; (2)~\textit{user intent}, an inference of the user's likely immediate goal; (3)~\textit{rationale}, a brief explanation of the proactive support decision grounded in the previous internal state and new screenshots; and (4)~\textit{$y_t$}, the binary support decision. Each $o_t$ is written to long-term memory $\mathcal{M}$ and consolidated into higher-level information about the user (\refsec{sec:memory}). Memory retrieval then constructs the context $h_t$ used at the next timestep, closing the personalized context-update loop:
\begin{equation}
h_t = \operatorname{Retrieve}(\mathcal{M}, o_t, t).
\end{equation}
When $y_t=1$, the Execution Agent receives $h_t$ as initial personalized context and may retrieve additional information from $\mathcal{M}$ before generating assistance. The user's subsequent reaction provides implicit feedback $r_t$, which is used to continually refine $\mathcal{O}$ to personalize its behavior (\refsec{sec:training}).

\subsection{Context Management and Long-Term User Modeling}
\label{sec:memory}

Continuous observation provides Coco with rich evidence about a user's ongoing activities, but retaining the complete observation stream in the Observer Agent's context is neither efficient nor necessary. Coco therefore maintains a long-term memory $\mathcal{M}$ that organizes the Observer Agent's textual outputs and extracts higher-level patterns about the user.


\noindent
\textbf{Proposition-Observation Bipartite Graph}\quad
We organize $\mathcal{M}$ as a bipartite graph with two sets of vertices: timestamped Observer outputs $\{o_1,\ldots,o_T\}$ and higher-level user propositions $\{p_1,\ldots,p_N\}$. We derive the latter using the General User Model (GUM)~\citep{shaikh2025creating}. Using the same model that powers $\mathcal{O}$, GUM periodically synthesizes propositions from sequences of Observer Agent outputs to capture recurring patterns in the user's activities. Each proposition $p_i$ represents an inferred property of the user, while edges connect $p_i$ to the observations that provide evidence for it. When a new proposition is synthesized, Coco retrieves the top-$k$ most similar existing propositions and determines whether the new evidence should create a new proposition or revise an existing one. When a proposition is revised, its supporting observation set is updated accordingly.

\textbf{Context Management}\quad
Memory serves two roles: maintaining compressed hidden state for the Observer Agent and providing long-term context to the Execution Agent. To retrieve content from $\mathcal{M}$, we focus on two logical primitives: temporal retrieval and semantic retrieval. After producing $o_t$, Coco retrieves the $k$ most recent Observer outputs together with the $k$ propositions most semantically relevant to $o_t$. Their combination forms the accumulated context $h_t$. 
$\mathcal{M}$ also bridges the Observer Agent and the Execution Agent invoked on demand. When $y_t=1$, Coco provides $h_t$ as initial context for proactive assistance and exposes $\mathcal{M}$ as a searchable tool that the Execution Agent can query. Rather than relying on a fixed retrieval policy or requiring the Observer Agent to determine all relevant context in advance, the Execution Agent can decide  what information is needed and retrieve it from $\mathcal{M}$ through multi-turn tool calls. This separation resembles the interaction between a file system and an agent that selectively searches it in modern coding-agent systems.

\subsection{Continual Adaptation via Self-Reflective Feedback}
\label{sec:training}

Long-term memory personalizes the context available to the Observer, but does not by itself change how $\mathcal{O}$ interprets that context or decides when to intervene. Coco therefore continually adapts the Observer Agent using implicit feedback derived from the user's responses to proactive suggestions and subsequent computer-use behavior.

\noindent
\textbf{Identifying Implicit Feedback}\quad
For timesteps at which Coco intervenes ($y_t=1$), we derive feedback $r_t$ from the user's subsequent response. We assign $r_t=1$ when the user engages with the proactive assistance and $r_t=0$ when the user rejects it, providing direct evidence about whether the intervention was appropriate. However, relying solely on these signals risks a degenerate policy in which $\mathcal{O}$ learns to intervene less frequently and receives fewer signals. To recover such missed opportunities, we employ the same on-device model in $\mathcal{O}$ to retrospectively scan $\{h_1, \ldots, h_T\}$ and identify cases in which the user subsequently exhibits evidence of needing assistance based on heuristics, such as repeatedly performing the same operation, becoming stuck, or explicitly invoking AI (see full prompts in Appendix~\ref{appendix:adaptation_data_prep_prompt}). When such evidence follows a timestep with $y_t=0$, we treat the preceding decision as a missed intervention opportunity and assign negative feedback to that decision. We acknowledge that this secondary signal identifies potential support opportunities rather than confirmed user preferences; nonetheless, it provides the exploration signal and subsequent user feedback under similar circumstances can further calibrate it.

\noindent
\textbf{Observer Adaptation}\quad
Once we infer $r_t$, we recover the corresponding Observer Agent input $(s_t, h_{t-1})$ and output $o_t$, and use the on-device model for self-reflection. When $r_t = 0$, the model produces a revised output $o_t'$; when $r_t = 1$, we set $o_t' = o_t$. The resulting dataset $\mathcal{D} = \{(s_t, h_{t-1}, o_t')\}$ in a certain period of time is used to update $\mathcal{O}$. While Coco is agnostic to the particular learning algorithm applied to $\mathcal{D}$, we instantiate two low-resource adaptation strategies: (1)~\textbf{LoRA Training (Coco-LoRA)}: examples in $\mathcal{D}$ are periodically shuffled and used to update the LoRA adapter of the multimodal language model that powers $\mathcal{O}$~\citep{hu2022lora}. (2)~\textbf{Prompt Evolution (Coco-PE):} following ACE~\citep{zhang2026agentic}, we maintain a compact bullet-point instruction memory that is iteratively refined through generation, reflection, and curation. At each batch, the current memory is used to generate predictions over $\mathcal{D}$. Incorrectly predicted examples are selected and augmented with temporally adjacent correctly predicted examples to provide contrastive evidence. These examples are then passed through reflection and curation to identify useful behavioral rules and update the Observer Agent prompt. Appendix~\ref{appendix:coco_pe_prompts} provides the details.

\section{Efficient On-device Deployment}
\label{sec:deployment}
\begin{figure*}[t]
  \centering
  \includegraphics[width=1\textwidth]{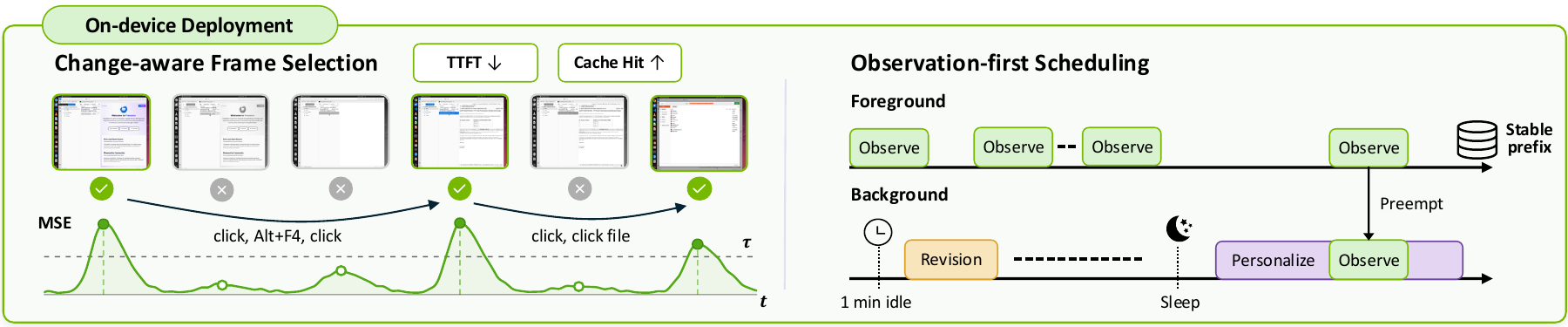}
\caption{\textbf{Efficient on-device deployment of the Observer Agent.} Coco uses change-aware frame deduplication to reduce time to first token (TTFT), together with workload-aware scheduling that prioritizes observation requests and performs personalization updates during idle periods.
}
\label{fig:deployment}
\end{figure*}

To preserve user privacy, Coco's Observer Agent preferably runs on-device, ensuring raw screen content never leaves the user's machine. Unlike conventional LLM serving in data centers, the Observer Agent faces a distinct workload: it must process high-frequency screen captures in real time and concurrently handle personalization updates. As illustrated in \reffig{fig:deployment}, we address these challenges through frame deduplication to reduce redundant token computation and priority scheduling.

\textbf{Change-aware Frame Selection}\quad
Consecutive screen captures are often nearly identical, particularly during typing and scrolling. However, even small visual differences can lead to different multimodal tokens due to the way multimodal language models encode images. Given Observer Agent input $s_t$, rather than feeding all screenshots to the multimodal model, we first align the user mouse and keyboard actions with each screenshot. Then we retain the initial and final frames but keeps an intermediate frame only when its RGB mean-squared error from the most recently retained frame exceeds a threshold $\tau$. Actions associated with discarded frames are forwarded to the next retained frame, preserving interaction semantics while avoiding redundant visual tokens. On a workload replayed on DGX Spark, this strategy raises the KV cache hit rate from 26.7\% to 45.2\% by removing redundant visual tokens that would otherwise evict reusable cache entries. This is especially important for on-device deployment under memory constraints.


\textbf{Observation-first Scheduling}\quad
Coco treats continuous observation and user-facing inference as foreground workloads, while personalization runs opportunistically in the background. Lightweight revision jobs become eligible after one minute of user inactivity, whereas full personalization is deferred until a longer idle period or an explicit sleep mode where the user turns off the continuous observation. Personalization executes as a bounded, checkpointed, low-priority process and is preempted whenever interactive inference requires the model. Besides preventing personalization from delaying observation, separating these heterogeneous workloads improves cache locality: consecutive Observer requests share a stable prompt prefix, allowing their KV-cache entries to remain resident instead of being frequently displaced by personalization prompts. Preempted jobs later resume from their latest checkpoint, enabling continual adaptation without compromising the responsiveness of the always-on Observer.

In real-world deployment, we combine these strategies with speculative decoding through block diffusion prediction~\citep{chen2026dflash} to further speed up inference. Appendix~\ref{appendix:profiling} provides the on-device deployment details and workload profiling statistics.

\section{Experiment}

\subsection{OSWorld-Persona: Repurposing Computer-Use Data for Proactive Agent}
\label{sec:dataset}

\begin{wrapfigure}{r}{0.65\textwidth}
  \centering
  \vspace{-1em}
  \includegraphics[width=\linewidth]{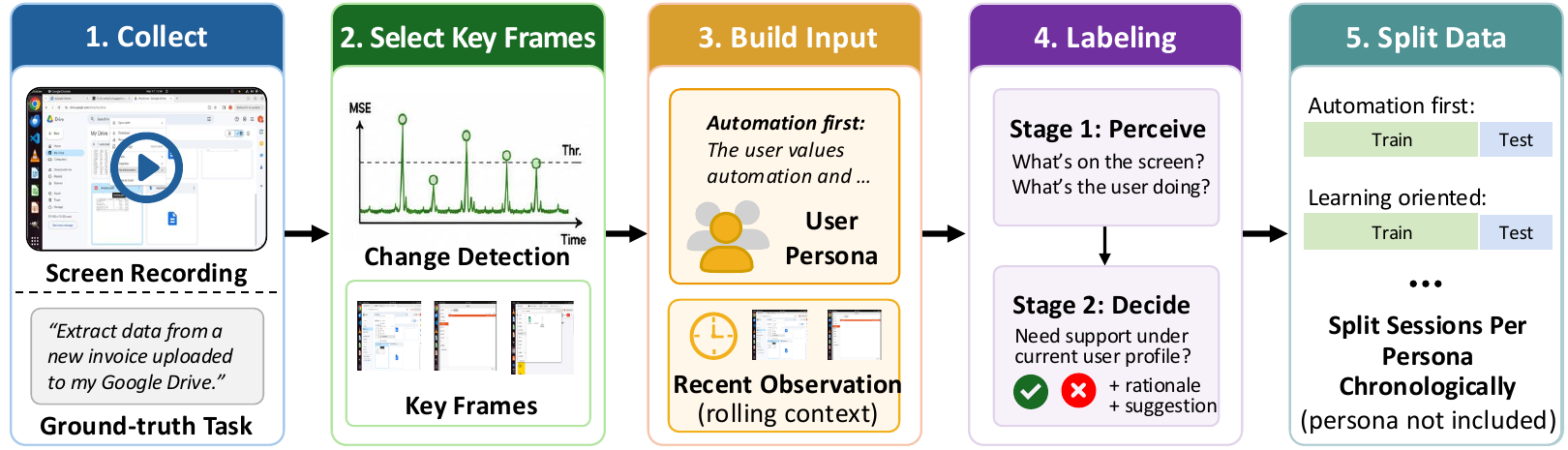}
  \caption{\textbf{Synthetic data pipeline for personalized proactive agent studies.}
  We use this pipeline to construct OSWorld-Persona from OSWorld v2 trajectories.}
  \label{fig:data_pipeline}
\end{wrapfigure}

Existing datasets for studying proactive agents are largely confined to text-only environments and do not capture the complexity of real-world workflows. To bridge this gap, we propose a general data synthesis pipeline that can leverage any computer-use dataset. The pipeline consists of five stages: First, we collect screen recordings from a computer-use dataset of choice; second, from these recordings, we follow \refsec{sec:deployment} to extract key frames that capture meaningful state transitions; third, we build input instances by pairing key frames in a rolling context with different user personas that describes the user's preferences and behavior patterns; fourth, the combined inputs are feed into a two-stage labeling process, where a VLM leverages the persona description to generate personalized assistance labels; finally, the labeled sessions are split chronologically into train and test sets per user. 


We instantiate this pipeline on OSWorld 2.0~\citep{yuan2026osworld2}, as it particularly curates long-horizon tasks that closely resemble everyday computer activities in both personal and professional scenarios. We consider 4 different personas ranging from wanting more support from AI (``automation first'') to only preferring proactive support when stuck (``learning oriented''), and employ Kimi-K2.6 to simulate the user's decision based on the given persona. For quality control, we apply a self-consistency check at temperature 0.4, retaining an example only when two additional runs reproduce the original \texttt{need\_support} label, and apply human validation by letting human to review cases where \texttt{need\_support} is true and annotating context that informs potential proactive support.  This yields \textbf{OSWorld-Persona}, comprising 6,558 key frames and 3,279 proactive support annotations from 108 sessions. Appendix~\ref{appendix:dataset} includes data synthesis prompts and detailed statistics.

\subsection{Evaluation Setup}

\noindent
\textbf{Models}\quad
We consider the following multimodal models that can be deployed locally on consumer hardware or AI PCs to cover different model architectures: (1) Qwen-3.5-9B, dense VLM with 9B activation parameters, (2) Qwen-3.5-35B-A3B, mixture-of-experts VLM with 3B activation parameters, (3) Nemotron-3-Nano-Omni, mixture-of-experts omni models with 3B activation parameters.

\noindent
\textbf{Baselines}\quad
We compare Coco observer adaptation method with (1) Vanilla, that prompts the off-the-shelf model to make the binary support decision; (2) Induced Persona, that induces compact user profile based on the training data~\citep{zhang-2024-guided}; (3) Oracle Persona, that includes the oracle persona used in data synthesis (not available in actual deployment). We further compare Coco context management with (1) relying on the model context and (2) Mem0~\citep{chhikara2025mem0} that uses LLM to extract salient information from observations and retrieve similar memories based on vector embeddings. For both Coco and Mem0, we use Qwen-3.5-9B to construct the memory, and use the answer model to decide the retrieval queries. We limit the number of retrieval call to 5.

\noindent
\textbf{Datasets and Metrics}\quad
We use \textbf{OSWorld-Persona} constructed in~\refsec{sec:dataset} to evaluate the proactive support quality. Following~\citet{lu2025proactive}, we report: (1) \textbf{Recall}, the fraction of genuine needs the agent identifies; (2) \textbf{Precision}, the fraction of offered interventions the user accepts; (3) \textbf{Accuracy}, overall correctness of the agent's predictions; (4) \textbf{False-Alarm}, the fraction of interventions offered when none was needed; and (5) \textbf{F1-Score} provides a balanced measure of the effective of the agent’s proactive behavior. To evaluate our context management strategy, we adopt standard memory benchmarks including \textbf{LoCoMo}~\citep{maharana2024evaluating}, \textbf{LongMemEval}~\citep{wu2025longmemeval}, \textbf{BEAM-100K}, and \textbf{BEAM-1M}~\citep{tavakoli2026beyond}. These benchmarks evaluate context management quality on a large number of conversation histories and use \textbf{LLM-as-a-judge Accuracy}. 
We use \texttt{gpt-5.5} as the judge model and follow the same judging protocol in \citet{chhikara2025mem0}\footnote{\url{https://github.com/mem0ai/memory-benchmarks}}.

\section{Results and Analysis}

\subsection{Main Results}

\begin{table}[t]
\centering
\resizebox{\textwidth}{!}{%
\begin{tabular}{llccccc cccc}
\toprule
\multirow{2}{*}{\textbf{Model}}
& \multirow{2}{*}{\textbf{Method}}
& \multicolumn{5}{c}{\textbf{Macro Average}}
& \multicolumn{4}{c}{\textbf{F1-Score Per Persona $\uparrow$}} \\
\cmidrule(lr){3-7}
\cmidrule(lr){8-11}
&
& \textbf{Precision $\uparrow$}
& \textbf{Recall $\uparrow$}
& \textbf{Acc. $\uparrow$}
& \textbf{FA $\downarrow$}
& \textbf{F1 $\uparrow$}
& \makecell{\textbf{Automation}\\\textbf{First}}
& \makecell{\textbf{Eager}\\\textbf{Hands-on}}
& \makecell{\textbf{Learning-}\\\textbf{oriented}}
& \makecell{\textbf{Reserved}\\\textbf{Draft}} \\
\midrule

\multirow{5}{*}{Qwen-3.5-9B}
& Vanilla
& 0.4231 & 0.1739 & 0.6498 & \textbf{0.0468} & 0.1861
& 0.0371 & 0.1387 & 0.3919 & 0.1765 \\

& Induced Persona
& 0.3729 & 0.2221 & 0.6380 & 0.1055 & 0.2094
& 0.0785 & 0.1993 & 0.2583 & 0.3014 \\

& Oracle Persona
& 0.4758 & 0.4345 & 0.7295 & 0.1599 & 0.3932
& 0.2096 & 0.7291 & 0.3099 & \textbf{0.3243} \\

& \textbf{Coco-PE}
& 0.4845 & \textbf{0.4933} & \textbf{0.7759} & 0.2492 & \textbf{0.4832}
& \textbf{0.5233} & 0.8088 & 0.3888 & 0.2121 \\

& \textbf{Coco-LoRA}
& \textbf{0.5052} & 0.4334 & 0.7753 & 0.2343 & 0.4551
& 0.3126 & \textbf{0.8408} & \textbf{0.4264} & 0.2404 \\

\midrule

\multirow{5}{*}{\makecell[l]{Qwen-3.5-\\35B-A3B}}
& Vanilla
& 0.3884 & 0.1863 & 0.6350 & \textbf{0.0857} & 0.1827
& 0.1020 & 0.1507 & 0.1900 & \textbf{0.2878} \\

& Induced Persona
& 0.4373 & 0.2886 & 0.6490 & 0.0864 & 0.2408
& 0.1184 & 0.2158 & \textbf{0.3440} & 0.2850 \\

& Oracle Persona
& 0.4448 & 0.3192 & 0.6527 & 0.1188 & 0.2713
& 0.2671 & 0.3810 & 0.1769 & 0.2602 \\

& \textbf{Coco-PE}
& \textbf{0.5678} & \textbf{0.3785} & \textbf{0.7846} & 0.1298 & \textbf{0.4334}
& \textbf{0.4514} & \textbf{0.7369} & 0.2603 & 0.2850 \\

& \textbf{Coco-LoRA}
& 0.4087 & 0.3335 & 0.6784 & 0.1697 & 0.3135
& 0.1911 & 0.5385 & 0.2563 & 0.2682 \\

\midrule

\multirow{5}{*}{\makecell[l]{Nemotron-3-\\Nano-Omni}}
& Vanilla
& 0.4050 & 0.0815 & 0.6485 & \textbf{0.0359} & 0.1175
& 0.0362 & 0.0951 & 0.1525 & 0.1863 \\

& Induced Persona
& 0.4260 & 0.1965 & 0.6494 & 0.0714 & 0.2192
& 0.0657 & 0.1991 & \textbf{0.3043} & \textbf{0.3075} \\

& Oracle Persona
& 0.4222 & 0.2218 & 0.6927 & 0.1285 & 0.2810
& 0.1721 & 0.5247 & 0.1515 & 0.2759 \\

& \textbf{Coco-PE}
& 0.4784 & 0.3025 & 0.6922 & 0.1233 & 0.3211
& \textbf{0.3229} & 0.4737 & 0.1928 & 0.2949 \\

& \textbf{Coco-LoRA}
& \textbf{0.4923} & \textbf{0.3152} & \textbf{0.7857} & 0.2237 & \textbf{0.3485}
& 0.1771 & \textbf{0.8609} & 0.1635 & 0.1923 \\

\bottomrule
\end{tabular}
}
\vspace{-0.5em}
\caption{\textbf{Results on OSWorld-Persona.} 
Compared to baselines, Coco-PE and Coco-LoRA achieve higher F1 scores across models and personas, but exhibit higher False-Alarm (FA) rates, as vanilla models rarely trigger proactive support. Results are averaged over three random seeds.}
\label{table:synthetic_main_results}
\end{table}

\textbf{Observer Adaptation Effectiveness}\quad
As shown in \reftab{table:synthetic_main_results}, Vanilla observers rarely identify support opportunities, yielding macro recall of only 0.082-0.186 and F1 scores below 0.19. Providing persona information alone improves performance, but remains insufficient: Induced Persona reaches F1 scores of 0.209-0.241, while Oracle Persona reaches 0.271-0.393. In contrast, Coco's adaptation strategies raise the best macro F1 to 0.483 on Qwen-3.5-9B (2.6$\times$ over vanilla), 0.433 on Qwen-3.5-35B-A3B (2.4$\times$), and 0.349 on Nemotron-3-Nano-Omni (3.0$\times$), even outperforming the corresponding Oracle Persona baseline. 
Crucially, these gains are not merely a byproduct of intervening more often as both precision and accuracy improve. 
The false alarm rate of the adapted observer is slightly higher than the vanilla observer as the adapted observer shift from near-silence toward active intervention. Coco-PE is the stronger method on the Qwen models, whereas Coco-LoRA is better on Nemotron-3-Nano-Omni.





\begin{wraptable}{r}{0.55\textwidth}
\centering
\vspace{-1em}

\resizebox{\linewidth}{!}{%
    \begin{tabular}{llccc}
\toprule
\textbf{Method} & \textbf{Variant}
& \makecell{\textbf{Qwen-3.5-}\\\textbf{9B}}
& \makecell{\textbf{Qwen-3.5-}\\\textbf{35B-A3B}}
& \makecell{\textbf{Nemotron-3-}\\\textbf{Nano-Omni}} \\
\midrule
\multirow{3}{*}{Coco-PE}
& Full & \textbf{0.4832} & \textbf{0.4334} & \textbf{0.3211} \\
& No Context & 0.3878 & 0.3983 & 0.3040 \\
& Decision Only & 0.3045 & 0.1993 & 0.2529 \\
\midrule
\multirow{3}{*}{Coco-LoRA}
& Full & \textbf{0.4551} & \textbf{0.3135} & \textbf{0.3485} \\
& No Context & 0.2851 & 0.2951 & 0.3247 \\
& Decision Only & 0.4286 & 0.3204 & 0.2450 \\
\bottomrule
\end{tabular}
}
\caption{\textbf{F1 scores of different Observer Agent formulations.}
The full Observer Agent achieves the best F1 under both Coco-PE and Coco-LoRA.}
\label{table:ablation}
\vspace{-1em}

\end{wraptable}

\textbf{Context Management Effectiveness}\quad
We ablated context management component in the Observation Agent by comparing against two restricted variants: (1) \textbf{Coco (No Context)}, which removes the accumulated activity context ($h_t$) from the Observer Agent input entirely, and (2) \textbf{Coco (Decision Only)}, which lets the Observer Agent to output the intervention decision alone rather than the structured output including the observation and user intent. Results in \reftab{table:ablation} show that the richer signal provided by context management is important for personalization adaptation. Concretely, under Coco-PE, removing the accumulated context reduces F1 from 0.483 to 0.388 on Qwen-3.5-9B, from 0.433 to 0.398 on Qwen-3.5-35B-A3B, and from 0.3211 to 0.3040 on Nemotron-3-Nano-Omni-30B-A3B. Under Coco-LoRA, removing context also substantially reduces F1, particularly on Qwen-3.5-9B, where F1 drops from 0.455 to 0.285. These results indicate that accumulated context is important for personalized proactive assistance and is not merely a prompting artifact. At the same time, by comparing against the results of Coco (Decision Only), we find letting the Observer Agent output the observation not only allows it to accumulate context but also helps the prediction of the proactive decision.

\begin{wraptable}{r}{0.55\textwidth}
\centering
\resizebox{\linewidth}{!}{%
    \begin{tabular}{llcccc}
\toprule
\makecell[l]{\textbf{Answer}\\\textbf{Model}} &
\textbf{Memory} &
\textbf{LoCoMo} &
\textbf{LongMemEval} &
\makecell{\textbf{BEAM}\\\textbf{-100K}} &
\makecell{\textbf{BEAM}\\\textbf{-1M}} \\
\midrule

\multirow{3}{*}{GPT-5.5}
& /
& 0.9364 & \textbf{0.9440} & 0.5811 & 0.5121 \\
& Mem0
& 0.7734 & 0.7220 & 0.6750 & 0.6257 \\
& \textbf{Coco}
& \textbf{0.9513} & 0.8220 & \textbf{0.7500} & \textbf{0.6786} \\
\midrule

\multirow{3}{*}{\makecell[l]{Claude-\\Sonnet-4.6}}
& /
& 0.9162 & \textbf{0.9040} & 0.5394 & 0.3840 \\
& Mem0
& 0.7240 & 0.7180 & 0.7275 & 0.7171 \\
& \textbf{Coco}
& \textbf{0.9448} & 0.7520 & \textbf{0.8375} & \textbf{0.8329} \\
\midrule

\multirow{3}{*}{Kimi-K3}
& /
& 0.9227 & \textbf{0.9200} & 0.6010 & 0.3951 \\
& Mem0
& 0.7955 & 0.7260 & 0.5966 & 0.5799 \\
& \textbf{Coco}
& \textbf{0.9565} & 0.8340 & \textbf{0.7194} & \textbf{0.7253} \\
\bottomrule
\end{tabular}
}
\caption{\textbf{Memory benchmark results.}
For Mem0 and Coco, Qwen-3.5-9B constructs the memory.}
\vspace{-1em}
\label{table:memory_results}
\end{wraptable}

We further compare the Coco's context management strategy with other alternative approaches on standard memory benchmarks. As shown in \reftab{table:memory_results}, across three different answer models, Coco consistently outperforms Mem0 on every benchmark and achieves the best results on LoCoMo, BEAM-100K, and BEAM-1M. The advantage is most pronounced on the longer BEAM settings: for example, with Claude Sonnet 4.6, Coco improves accuracy from 0.539 to 0.838 on BEAM-100K and from 0.384 to 0.833 on BEAM-1M relative to using no external memory. Coco similarly raises BEAM-1M performance from 0.512 to 0.679 with GPT-5.5 and from 0.395 to 0.725 with Kimi-K3. The vanilla long-context models perform best on LongMemEval as conversations in this benchmark fit within the context windows; but it does work well when the memory goes longer.

\subsection{User Study}
\label{sec:user_study}

\begin{figure}[t]
    \centering
    \includegraphics[width=\linewidth]{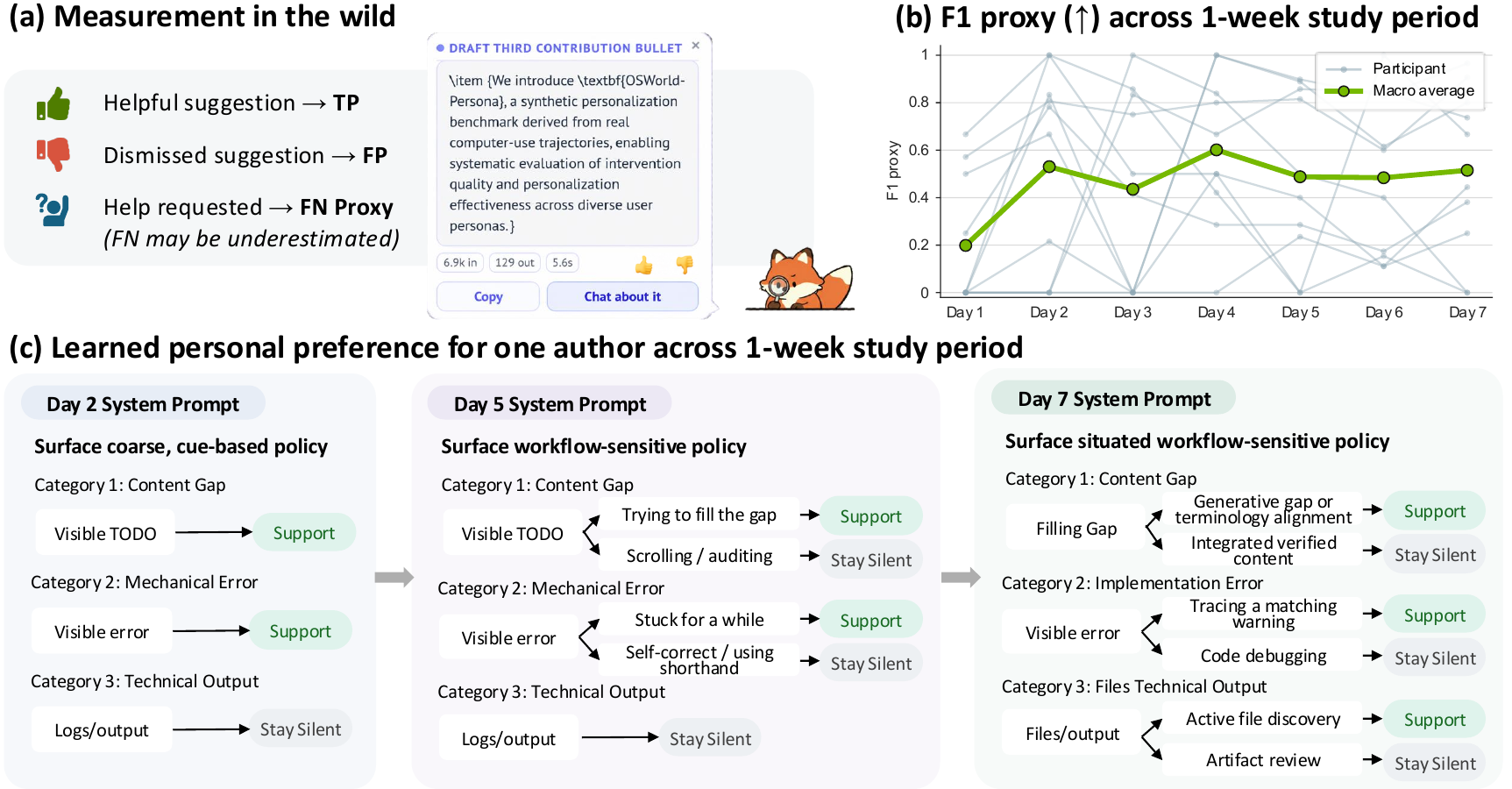}
    \caption{\textbf{Deep personalization in the wild.} In a week-long study with 10 participants who used Coco during their everyday computer activities, Coco’s macro F1 proxy rose from 0.199 to 0.515. The learned boundaries through deep personalization shows a trend of progressive refinement, from surfacing coarse, cue-based policy to surfacing situated workflow-sensitive policy.
    }
    \label{fig:user_study_results}
\end{figure}

To assess whether Coco's on-device deep personalization yields useful and well-timed assistance in real-world settings, we instantiate Coco as a desktop assistant and conduct a user study in which 10 participants used Coco during their everyday computer activities for one week. The Observer Agent runs Qwen3.5-9B on a single RTX 5090 GPU, and is adapted using Coco-PE, the best-performing method for this backbone according to our benchmark results. During the study, we explain Coco's proactive functionality to participants and provide minimal instructions, allowing them to use their computers as normal. The only instructions are to use the thumb-up/thumb-down buttons and to request help when desired through Coco's text or audio interface. Appendix~\ref{appendix:coco_interface} provides further details on Coco desktop assistant implementation.

The Observer Agent's inputs, outputs, and curated memory remain on the user's device. With informed consent, we collect only three types of interaction events as illustrated in \reffig{fig:user_study_results} (a): (1) a thumb-up on a proactive suggestion, treated as a true positive; (2) a thumb-down or dismissing the suggestion without revealing the full content, treated as a false positive; (3) a direct text or audio request when Coco remains silent, treated as a proxy for a false negative\footnote{False negatives may be underestimated, as users may desire assistance without explicitly prompting Coco.}. \reffig{fig:user_study_results} (b) reports the F1 proxy over the one-week study period. While fluctuations exist for individual users on certain days, the macro-averaged F1 proxy improves steadily over time, rising from 0.199 to 0.515 during the study period. The results also reflect a cold-start effect: on Day 1, 6 out of 10 users have an F1 proxy of zero because Coco produces no helpful proactive suggestions. After adaptation, all six begin receiving suggestions they rate as helpful.

Because participants' observation data remain private, we qualitatively examine an author's week-long deployment separately, reviewing the local adaptation data $\mathcal{D} = \{(s_t, h_{t-1}, o_t')\}$ and the resulting evolved system prompt. \reffig{fig:user_study_results} (c) illustrate three categories of learned decision boundaries on different days. Interestingly, the learned boundaries in the system prompt shows a trend of progressive refinement. By Day 2, the deep personalization procedure learns coarse cues, such as a visible ``TODO'' often signals an opportunity to help, whereas technical output often calls for silence. Subsequent feedback allows Coco to learn more nuanced understanding of the user's typical workflows. For example, on Day 5, Coco learns to assist when the user is actively filling a ``TODO'', but to remain silent when the user is scrolling through or auditing the full document. By Day 7, the prompt further distinguishes between filling a generative gap, where assistance is welcome, and integrating verified content, where the user prefers doing it themselves. These refinements demonstrate how deep personalization learns situated preferences beyond the static profiles or explicit settings used by conventional personalization approaches.


\section{Related Work}
\noindent
\textbf{Proactive Computing}\quad
Proactive computing enables systems to anticipate needs and act before explicit requests, which pushes the limit of computing and elicits new needs that can lead to increased human productivity and quality of life~\citep{tennenhouse2000proactive}. It has been applied to information retrieval~\citep{rhodes2000just}, recommendation~\citep{resnick1997recommender,sarwar2001item,covington2016deep},
programming assistance~\citep{pu2025assistance,liu2024repobench}, and ubiquitous computing~\citep{weiser1991computer}. 
In modern LLM systems, proactive agents remain nascent. \citet{lu2025proactive} formulate the proactive agent problem and define evaluation metrics, and subsequent work studies proactivity in mobile agent and conversational settings~\citep{yang2026fingertip,hu2026anticipate}. Our work focuses on personalization as proactive assistance is valuable only when it surfaces the right thing at the right moment, a judgment that is inherently subjective~\citep{horvitz1999principles}.

\noindent
\textbf{Data Source for Personalization}\quad
Personalization commonly draws on user messages, writing, and other digital traces~\citep{zhang2025personalization,salemi2024lamp}, although these signals can be noisy. More explicit approaches solicit pairwise preferences~\citep{li2024personalized}, demonstrations~\citep{shaikh2025aligning}, or direct edits to prompts and personas~\citep{muller2026personalizing}. A second family infers signal from natural interaction instead of soliciting it. For example, \citet{sharma2022correcting} adapt a robot planning policy from human corrections, and \citet{gao2024aligning} recover latent preferences from user edits to LLM agent output. These methods scale better than direct elicitation but often tied to a specific task. A third option infers preferences from user activity. In
\citet{shaikh2025creating}, general preference propositions are induced from raw observations of computer use, but these propositions are open-ended and are not confirmed by the user. Building on the last two directions, we propose a deep personalization framework that learns user preferences through continuous context observation.

\noindent
\textbf{Personalized Agent Adaptation}\quad
In agentic systems, personalization affects the full decision pipeline rather than only response generation~\citep{zhang2025personalization,xu2026toward}. Unlike adaptation with verifiable objectives and ground-truth evaluators~\citep{yuksekgonul2026learning,novikov2025alphaevolve}, personalized adaptation must learn latent, evolving preferences from a single user's feedback, often while keeping sensitive data on device. One line of work applies parameter-efficient fine-tuning to user data~\citep{tan2024democratizing,wang2026edgetune}, or specializes adapters for individual components of the agentic system such as memory~\citep{bini2025memlora}. With the growing instruction-following capabilities of LLMs, recent work also induces personalized patterns into retrievable skills~\citep{xia2026metaclaw}. While prior work in agentic settings largely rely on large cloud-based models, we show that on-device-scale models can also provides a knob for agent personalization.


\section{Conclusion}
We present Coco, a personalized proactive agent that realizes deep personalization by closing the loop of continuous context observation, long-term user modeling and self-adaptation from implicit user feedback with an Observer Agent. By efficiently deploying the Observer Agent locally, we further demonstrate that privacy-preserving personalization is not only principled but practical on a single consumer device. To support rigorous evaluation of personalized proactive agents, we introduce OSWorld-Persona, a new synthetic benchmark constructed via data synthesis pipeline that enables evaluation of deep personalization without releasing real user data. Experiments on OSWorld-Persona and a week-long user study show that personalization is essential for proactive support, and that Coco is able to adaptive to individual needs in the wild.

\section*{Acknowledgments}
We are grateful for discussions and help from Shang Yang, Ligeng Zhu, Han Cai, Zhiding Yu, Yanzhe Zhang, Zora Wang, Haowen Wang, Shunta Kimura, Fenghai Li, Minyang Xie, Weixian Xu, Xinrui Shi, and the broader NVIDIA community.

\section*{AI Use Statement}
In this work, LLMs are used to assist with code development and also used as components of the data synthesis pipeline for support-decision labeling. We manually check all model-generated content. LLMs are not used for research ideation, result interpretation, experimental design, or generating claims about system performance. We also use Coco, the local proactive assistant introduced in this work, to support daily research work and section drafting during the writing process. We take responsibility for the final content of this work, including text, claims, and artifacts produced with the aid of generative AI.

\section*{Ethics Statement}
This paper proposes deep personalization, a closed-loop process connecting continuous context observation with continual agent adaptation. We instantiate this framework as Coco, a proactive agent that improves suggestion timing and reducing disruptive interruptions. We see personalized proactive agents holding particular promise for democratizing AI assistance: many people do not yet recognize when AI could help them~\citep{savoldi2025generative,zhang2025mapping}, and a system that observes context and intervenes at the right moment can bridge the gap.

While our experiments demonstrate its effectiveness, they also surface an inherent tension fundamental to personalized AI research more broadly: the more user context and feedback an agent can observe and retain, the better it personalizes; yet broader observation increases exposure to sensitive information and risks revealing personal habits and preferences to malicious parties. Our deep personalization approach directly addresses this tension. By separating the Observer Agent from the broader system, we confine continuous desktop observation and adaptation to a self-contained, on-device component. We further contribute efficient on-device deployment strategies and demonstrate that real users can be served using local compute, including a single consumer RTX 5090 GPU or DGX Spark. Although the deep personalization algorithm is not tied to any particular system instantiation, we view on-device deployment as more than an engineering optimization, but as a principled privacy safeguard. By keeping observation, context retention, and adaptation local, the system reduces the need to route sensitive user data through third-party providers. We hope this encourages future work on personalized agents to prioritize privacy-preserving, on-device adaptation, enabling agents to better align with individual needs and preferences while minimizing unnecessary data exposure.

We nonetheless acknowledge several residual risks. First, continual adaptation may reinforce undesirable behavioral patterns or overfit to transient user behavior. Practical deployments should therefore provide user-accessible controls to inspect, edit, correct, or reset learned context. Second, the same observational capability that enables personalization could be misused as a surveillance mechanism if deployed without appropriate safeguards. We strongly caution that any deployment of such systems must require explicit, informed, and revocable user consent, with clear disclosure of what is observed, what is retained, and how the learned context and preference is used.

This paper does not release any user-activity dataset. Instead, we introduce a data synthesis pipeline that leverages existing computer-use datasets to study personalized proactive agents without exposing real user activity traces. Our one-week user study was conducted with full informed consent from all participants. During the study, we only collected aggregate interaction statistics, including the number of helpful proactive suggestions, unhelpful proactive suggestions, and user-requested assistance events. The Observer Agent was deployed locally using a self-hosted Qwen3.5-9B model on an RTX 5090 GPU, so participant desktop observations were processed on-device rather than sent to third-party model providers.

\bibliographystyle{unsrtnat}
\bibliography{ref}
  
\newpage
\appendix
\section{Limitations and Future Work}

While on-device deep personalization offers a new approach to adapting agents to individual users, and our evaluation of Coco demonstrates its potential for improving proactive assistance, several limitations remain.

\paragraph{Enhance User Control in Deep Personalization}
This work focuses on deep personalization as a closed-loop process that connects continuous context observation with continual agent adaptation. In Coco, much of the training signal is inferred implicitly from user behavior. Although participants appreciated that Coco could learn without requiring frequent explicit feedback, several also wanted greater visibility into what the system had learned and more opportunities to participate in the learning process. This points to an important challenge for future work: enabling users to inspect, correct, forget, or otherwise shape an agent's learned behavior without imposing substantial cognitive or interaction costs. Promising directions include lightweight explanations of why an intervention was suggested, editable representations of learned preferences, and controls that let users adjust the scope and pace of personalization.

\paragraph{Develop Proactive Agent that Augment Users}
Personalized proactive agents may help democratize access to AI assistance because users do not always recognize when or how an AI system could support them. An agent that understands the user's context and intervenes at an appropriate moment can help bridge this gap. At the same time, poorly designed proactive assistance could encourage users to trade understanding and independent thought for convenience. One participant in our study illustrates a more promising form of augmentation based on her own usage. After Coco suggested generating quiz questions while she was learning new material, she adopted the interaction repeatedly, and Coco became better calibrated to this behavior over the following days. Such interventions support the user's own learning process rather than simply completing the task on their behalf. Future work should investigate how agents can help users discover similarly beneficial forms of assistance, particularly during cold start, and how deep personalization can identify interventions that promote users' longer-term capabilities. This also calls for evaluation criteria that capture learning, agency, and sustained benefit in addition to immediate suggestion acceptance.

\paragraph{Privacy of Personalized Agent}
Personalized AI systems face an inherent tension: access to richer user context and feedback can improve personalization, but broader observation also increases exposure to sensitive information and the risk that personal activities, habits, or preferences are disclosed. Our architecture partially addresses this tension by separating the Observer Agent from the broader system and confining continuous desktop observation and adaptation to a self-contained, on-device component. However, Coco's execution agent currently benefits from larger models hosted in the cloud. Retrieved user context may therefore leave the device when the execution agent generates a response, meaning that our architecture reduces but does not eliminate privacy risks. We hope to explore fully on-device personalized agentic systems as local model capabilities and hardware efficiency improve.

\paragraph{Energy Consumption of Always-on Agent}
Continuous context sensing, inference, and adaptation introduce computational and energy costs that are absent from conventional, user-invoked assistants. These costs are particularly important on battery-powered devices, where an always-on agent may affect battery life, thermal performance, and the resources available to other applications. We have not yet systematically characterized these costs in this work. Future work should measure the energy-latency-quality trade-offs of deep personalization and explore more efficient designs, including event-triggered sensing, adaptive observation frequencies and model compression.
  
\section{Prompts Used in Coco Observer Agent}\quad

\subsection{Observer Agent Prompt}
We use the following prompt to turn a multimodal language model into a state machine that processes latest observations to update its own state and make proactive support decision.
\begin{promptbox}{Observer Agent Prompt}
\footnotesize
You are an OBSERVER in an everyday AI-support system. Your role is to analyze the user's screen activity and input to understand what they are doing, and to spot moments where a capable AI tool or agent could take a task off their hands. You must focus solely on understanding, describing, and inferring.
You will receive the following input blocks:\\

$<$memory$>$\\
Long-term personalized context about this user — preferences, the AI tools they have, approaches or guidance that worked for them before, and recurring tasks — accumulated across sessions. Often "(no memory yet)" when nothing has been learned. When present, use it to tailor your analysis to this specific user; it is NOT a description of the current screen.\\
$<$/memory$>$\\

$<$screenshots$>$\\
Historical images associated with user actions or meaningful screen changes. Each image corresponds to a timestamp and, when available, a monitor ID listed in the text. Action-triggered sequences begin with a before frame and end with an after frame; visually redundant intermediate frames may be removed. Every keyboard, click, or scroll action belonging to a removed frame is carried forward and listed on the next retained screenshot entry. A screenshot can therefore list multiple actions. Treat those actions as ordered direct evidence of what the user did before reaching the associated UI state. Compare chronology within each monitor; do not assume images from different monitors are successive states of one screen.\\
$<$/screenshots$>$\\

$<$recent\_observations$>$\\
Your last few observations and how the user reacted to the bubble each one triggered (ACCEPTED / DISMISSED / NEGATIVE rating / ignored / no feedback). May be absent at the start of a session.\\
$<$/recent\_observations$>$\\

$<$recent\_propositions$>$\\
Broader summaries of recent user activities inferred by long-term memory. These cover a longer horizon than recent\_observations and may describe an earlier task or workflow.\\
$<$/recent\_propositions$>$\\

$<$relevant\_propositions$>$\\
Up to three long-term memory summaries relevance-ranked from the explicit user input and newest prior observer state. These are distinct from recent\_propositions and can be older. They are retrieval hints, not evidence of the current screen.\\
$<$/relevant\_propositions$>$\\

Your responsibilities:
\begin{itemize}
    \item {Understand the timeline of activity. If you are given multiple screenshots with timestamps, treat them as a continuous workflow. Analyze (1) how the screen content changed over time; (2) what applications, websites, or documents the user is working in; (3) whether the user is switching between tools or staying in one place; (4) which monitor each historical action occurred on.}
    \item {Describe the current screen state.
Summarize the relevant content visible across the current monitor-reference group. Do not treat the final image as the only current screen. Reference earlier historical screenshots to clarify how each monitor's state evolved, if available.}
    \item {Infer the user’s intention. Based on the screenshots and the user’s input, infer what task or subtask the user is trying to accomplish right now.}
    \item {Detect support opportunities.
Use ``yes'' only when proactive support is likely to be genuinely useful at this moment and the benefit of interruption clearly outweighs the cost.}
\end{itemize}

Output in JSON format:
\begin{verbatim}
{
  "observation": "<compact incremental memory update>",
  "user_intent": "<the user's likely immediate goal>",
  "rationale": "<brief explanation of the need_support decision>",
  "need_support": "yes" | "no"
}
\end{verbatim}
\end{promptbox}

\subsection{Context Management and Long-Term User Modeling Prompts}
We adapt prompts from the General User Model~\citep{shaikh2025creating} to synthesize propositions from sequences of Observer Agent outputs. Coco maintains the proposition-observation bipartite graph to update Observer Agent hidden state and pass content to the Execution Agent when requested.

\begin{promptbox}{Proposition Synthesis Prompt for Context Management}
\footnotesize
You maintain a grounded model of a user's work context.
Infer only claims supported by the supplied observations. Preserve named tools, documents, projects, people, and tasks because they improve later retrieval.
Return JSON only.\\

Analyze these observations about \{user\_name\}:

\{observations\}

Every observation begins with a stable ``[observation\_id=...]`` marker. Return up to {max\_propositions} non-overlapping propositions as:
\begin{verbatim}
{{"propositions":[{{"proposition":"...","reasoning":"...",
"confidence":1,"decay":1,"observation_ids":["id"]}}]}}    
\end{verbatim}

Each proposition must cite only the observation IDs that directly support it. Confidence is 1-10 evidence strength. Decay is 1 for short-lived context and 10 for durable context. Be conservative and do not treat visible content as proof
of a lasting preference.
\end{promptbox}

\begin{promptbox}{New Proposition Classification Prompt for Context Management}
\footnotesize
Classify how a new proposition relates to existing ones.
Return JSON only. IDENTICAL means the same claim, SIMILAR means overlapping claims that should be consolidated, and UNRELATED means neither.\\

New proposition:

[new] \{new\_text\}

Reasoning: \{new\_reasoning\}

Existing propositions:

{existing}

Return
\begin{verbatim}
{{"label":"IDENTICAL|SIMILAR|UNRELATED","target_ids":[1,2]}}.  
\end{verbatim}
Only include IDs that directly participate in the stated relationship.
\end{promptbox}

\begin{promptbox}{Proposition Consolidation Prompt for Context Management}
\footnotesize
Consolidate a cluster of similar user-context propositions.
Produce a clear, non-redundant replacement set grounded only in the supplied observations. Preserve named people, applications, documents, projects, and tools when supported because they improve retrieval. Return JSON only.\\

Similar existing propositions:
\{propositions\}

New inferred claim:

\{new\_text\}

Reasoning: \{new\_reasoning\}

Supporting observations:

\{observations\}

Rewrite, merge, or split the similar claims into at most \{max\_propositions\} non-overlapping propositions. Every resulting proposition must cite only the observation IDs that directly support it. Return:
\begin{verbatim}
{{"propositions":[{{"proposition":"...","reasoning":"...",
"confidence":1,"decay":1,"observation_ids":["id"]}}]}}
\end{verbatim}
Confidence is 1-10 evidence strength. Decay is 1 for short-lived context and 10 for durable context.
\end{promptbox}

\subsection{Observer Adaptation Data Preparation Prompts}
\label{appendix:adaptation_data_prep_prompt}
As described in \refsec{sec:training}, the training data for Observer Agent adaptation comes from identifying implicit user feedback and revising the original Observer Agent output with self-reflection based on the signals. ``Prompt to Discover Missed Support Opportunities'' and ``Prompt to Ground Missed Support To Observation'' are the two prompts we used to retrospectively scan past observations and identify cases in which the user subsequently exhibits evidence of needing assistance. Together with the signals directly inferred from user's reactions to proactive suggestions, we use ``Self-reflection Prompt to Produce Revised Observation'' to obtain $o'_t$. The resulting dataset $\mathcal{D}=\{(s_t, h_{t-1}, o'_t)\}$ is used to update the Observer Agent $\mathcal{O}$. 

\begin{promptbox}{Prompt to Discover Missed Support Opportunities}
You retrospectively inspect a large chronological desktop-work timeline to discover a small set of HIGH-LEVEL, reusable proactive-support opportunities.\\

Use these retrospective principles:
\begin{itemize}
    \item {Later behavior may reveal repetitive work, stuckness, or an anticipatable need, but assistance must have been inferable before the later behavior.}
    \item {A later manual invocation of chatbot (e.g. ChatGPT), AI agent, or another assistance tool can be positive evidence that Coco missed an earlier opportunity to provide the same coherent support. Infer the opportunity at the earlier trigger; do not propose interrupting after equivalent assistance is already active.}
    \item {Do not treat every AI-tool use as a missed opportunity. It must reveal a reusable, high-value need that was inferable from the preceding workflow.}
    \item {Support must save substantial time, prevent a meaningful error, or unblock a task strongly enough to justify interruption.}
    \item {Respect stated user preferences and prior negative feedback. Never infer a preference merely from silence.}
    \item {Be conservative and use only supplied evidence.}
\end{itemize}

Reason across the WHOLE timeline instead of labeling each observation independently. Merge repeated local symptoms into the underlying workflow. Do not produce narrow opportunities such as fixing one SSH failure, correcting one command, flagging one typo, or explaining one transient error. Those events may be evidence of a broader workflow, but the proposed support must remain useful if the local symptom disappears.

At this stage, find the opportunity and its supporting evidence only.\\

Do NOT select a correction target or trigger. evidence\_observation\_ids must contain at least three UNIQUE supplied IDs establishing what the user later did or that the workflow recurred.\\

Evidence is behavioral proof, not a label target: an evidence observation may have original\_need\_support="yes" or "no" and does not need to be a candidate\_no\_intervention moment. Later manual AI use is valid evidence.\\
Prefer evidence distributed across time rather than several nearly identical adjacent screens. Describe when Coco should offer help, what coherent work it should do, and why the benefit clears the interruption cost. A later stage will inspect the entire chunk to find an earlier no-intervention trigger.

Return only this JSON object:
\begin{verbatim}
{
  "support_opportunities": [
    {
      "category": "repetitive_work" | "stuck" | "anticipate_need",
      "workflow_pattern": "<general recurring  user workflow>",
      "evidence_observation_ids": ["<at least 3 unique supplied IDs>"],
      "evidence_summary": "<how the cited observations establish the 
      pattern>",
      "when_to_offer": "If the user ..., offer ...",
      "why_high_value": "<why benefit clearly exceeds interruption 
      cost>",
      "why_not_incident_specific": "<why useful beyond one local 
      symptom>"
    }
  ]
}
\end{verbatim}

Return an empty list if no workflow-level opportunity is sufficiently supported.\\

Chronological desktop-work timeline:

\{list of observations\}
\end{promptbox}

\begin{promptbox}{Prompt to Ground Missed Support To Observation}
You ground verified workflow-level support opportunities to earlier missed intervention moments. Evidence discovery and verification have already finished for ALL chunks. You receive every verified opportunity with its evidence summary, IDs, and timestamps, plus one target chronological chunk whose observations include candidate\_no\_intervention. Label triggers only in that target chunk; use the verified evidence summaries and timing to judge what happened afterward.\\

Return exactly one decision for EVERY supplied verified opportunity. Never omit one. For an accepted decision:
\begin{enumerate}
    \item {Select trigger\_observation\_id from an observation whose candidate\_no\_intervention is true.}
    \item {Select the earliest moment where the verified broad need was already inferable and valuable enough to interrupt}
    \item {Later manual ChatGPT or agent use is positive evidence of an earlier need. Reject as redundant only if equivalent help was already active at the proposed trigger.}
    \item {Set confidence to at least 0.75 only if the trigger is clearly supported.}
    \item {Write rationale to explain why need\_support="yes" at the selected trigger. Ground it in what was observable at that trigger and the applicable support opportunity. Later evidence may disambiguate the need, but do not describe future behavior as if it were already visible.}
\end{enumerate}

Return only this JSON object:
\begin{verbatim}
{
  "trigger_decisions": [
    {
      "opportunity_id": "<exact supplied opportunity_id>",
      "accepted": true,
      "decision_rationale": "<specific reason for accepting/rejecting>",
      "trigger_observation_id": "<exact supplied eligible ID>",
      "rationale": "<why proactive support is needed at this trigger>",
      "confidence": <number between 0 and 1>
    }
  ]
}
\end{verbatim}
Rejected decisions need only opportunity\_id, accepted=false, and a concrete decision\_rationale. The list must contain exactly one decision per supplied verified opportunity.\\

Annotation payload:

\{a dict with "verified\_opportunities", "eligible\_trigger\_observation\_ids", "timeline"\}

\end{promptbox}

\begin{promptbox}{Self-reflection Prompt to Produce Revised Observation}
You correct observation annotations after user behavior has established that the original proactive-support prediction had the wrong polarity.\\

Treat the derived need\_support label as authoritative, but remain grounded in the provided record. Do not invent screen contents, actions, errors, or goals. If need\_support is "yes", clarify the concrete need or useful assistance opportunity supported by the record. If it is "no", remove unsupported claims of struggle, error, or interruption-worthy inefficiency. Do not mention labels, feedback, or the correction process in either field.\\

The revised observation and intent may be stored in long-term user memory. Preserve the concrete, memory-relevant details already visible in the original screenshot-derived observation, such as application and file names, commands, error messages, artifacts, workflow state, and what changed over time. Do not replace necessary screenshot details with a vague summary. Preserve only details supported by the record, and avoid incidental or sensitive details that are not needed to understand the user's work.\\

Return only a JSON object with exactly these string fields:
\begin{verbatim}
{
  "observation": "<revised factual description of the user's current
  situation>",
  "user_intent": "<the user's immediate goal, in under 15 words>"
}
\end{verbatim}

Original need\_support prediction: \{original\_prediction\}

Derived need\_support label: \{need\_support\_label\}

Derived label rationale: \{label\_rationale\}

Original observer input: \{observer\_input\}

Original observer output: \{observer\_output\}

\end{promptbox}

\subsection{Coco-PE Prompts}
\label{appendix:coco_pe_prompts}
We implement the prompt evolution approach following the Generate-Reflect-Curate procedure proposed in ACE~\citep{zhang2026agentic}. At each batch, the updated Observer Agent system prompt is used to generate predictions over $\mathcal{D}$. We use the ``Reflector Prompt'' below to reflects on the incorrectly predicted examples to produce insights about the user. These insights are passed to the ``Curator Prompt'' to identify useful behavioral rules and update the Observer Agent prompt. During the user study, we further use ``Final Organization Prompt'' to organize behavioral rules in the updated prompt into more human readable format with sections and categories. This allows the user to review what Coco learns through deep personalization.

\begin{promptbox}{Reflector Prompt}
You are the reflection module of a self-improving proactive desktop assistant.

The assistant maintains a MEMORY (a list of bullets, each with an id) describing one specific user's preferences for proactive support. On a training example the assistant made a prediction using that memory; you are given the prediction and the ground-truth label (what the ideal personalized assistant did, with its rationale).\\

Diagnose the gap and distill durable, REUSABLE lessons about this user:
\begin{itemize}
    \item {If the prediction was wrong, identify the root cause: which situation cue was misread, or which user preference was missing/misstated in the memory.}
    \item {If the prediction was right, note which memory bullets contributed so they can be reinforced. Only propose a new insight for a correct example if it captures a preference not yet in memory.}
    \item {Insights must be about the USER'S PREFERENCES and decision boundaries ("in situation X this user wants / does not want Y because Z"), not about this single example. Never mention example ids or screenshots.}
\end{itemize}

Respond with ONLY a JSON object:
\begin{verbatim}
{
  "reflection": "<1-3 sentences: what went wrong or right and why>",
  "helpful_bullet_ids": ["<ids of memory bullets that pointed the right 
  way>"],
  "harmful_bullet_ids": ["<ids of memory bullets that pointed the wrong
  way>"],
  "proposed_insights": [
    {"section": "when_to_support" | "when_to_stay_silent" | 
    "how_to_support" | "general",
     "content": "<one concrete, generalizable preference rule>"}
  ]
}
\end{verbatim}
Propose at most 2 insights; an empty list is fine.\\

$<$current\_memory$>$

\{memory\}

$<$/current\_memory$>$\\

$<$situation$>$

Ground-truth observation of the screen: \{gt\_observation\}

Ground-truth user intent: \{gt\_intent\}

$<$/situation$>$\\

$<$assistant\_prediction$>$

\{prediction\}

$<$/assistant\_prediction$>$\\

$<$ground\_truth\_label$>$

need\_support: \{gt\_need\}

suggestion\_type: \{gt\_stype\}

suggestion: \{gt\_suggestion\}

label rationale: \{gt\_rationale\}

$<$/ground\_truth\_label$>$\\

The prediction was \{verdict\}. Reflect and return only the JSON object.
\end{promptbox}

\begin{promptbox}{Curator Prompt}
You are the curator of the memory of a self-improving proactive desktop assistant. The memory is a sectioned list of bullets, each with an id, describing ONE user's preferences for proactive support.\\

You are given the current memory and a batch of reflections (lessons from recent training examples, each possibly proposing new insights). Update the memory with INCREMENTAL delta operations — do not rewrite it wholesale:
\begin{itemize}
    \item {"add":    a genuinely new preference rule not covered by any existing bullet.}
    \item {"update": rewrite an existing bullet to be more precise (e.g. sharpen its condition, merge in a nuance from a reflection).}
    \item {"delete": remove a bullet that reflections showed to be wrong or misleading.}
\end{itemize}

~
Rules:
\begin{itemize}
    \item {Keep bullets atomic, concrete and operational ("when X, do/don't Y"), one decision rule per bullet.}
    \item {Do NOT add duplicates or near-duplicates of existing bullets; prefer "update" to sharpen an existing bullet instead.}
    \item {Prefer few high-value operations over many marginal ones.}
\end{itemize}

Respond with ONLY a JSON object:
\begin{verbatim}
{
  "ops": [
    {"op": "add", "section": "when_to_support" | "when_to_stay_silent"
    | "how_to_support" | "general", "content": "..."},
    {"op": "update", "id": "<existing bullet id>", "content": "..."},
    {"op": "delete", "id": "<existing bullet id>"}
  ]
}
\end{verbatim}
At most \{max\_ops\} operations; an empty list is fine.\\

$<$current\_memory$>$
\{memory\}
$<$/current\_memory$>$\\

$<$reflections$>$
\{reflections\}
$<$/reflections$>$\\

Return only the JSON object with the delta operations.
\end{promptbox}

\begin{promptbox}{Final Organization Prompt}
You are the final inference stage of a self-improving proactive desktop assistant.

You are given an evolved memory containing detailed situation-level rules learned from one user's behavior. Convert that case-oriented memory into a compact model of the USER'S underlying intentions and assistance preferences.\\

Express each inference as one unified insight that connects what the user is trying to achieve with when or how they prefer proactive assistance. Do not create separate lists for intentions and preferences. An example is an existing detailed memory bullet that concretely supports an insight; refer to examples by bullet id instead of copying their text.\\

Rules:
\begin{itemize}
    \item {Infer across multiple bullets; do not merely paraphrase or enumerate them.}
    \item {Do not treat a recurring application, domain, or task as a preference.}
    \item {Keep each insight concise, distinct, and user-specific.}
    \item {Preserve important exceptions and decision boundaries in the insight itself.}
    \item {Choose a useful number of representative example bullet ids for each insight.}
    \item {Use each example only where it provides real evidence; not every source bullet needs to appear in the final memory.}
    \item {Decide how many insights and examples the evidence supports; do not target a predetermined count.}
    \item {Never invent an example id.}
\end{itemize}
~

Respond with ONLY a JSON object:
\begin{verbatim}
{
  "insights": [
    {
      "section": "when_to_support" | "when_to_stay_silent" 
      | "how_to_support" | "general",
      "content": "<unified insight about user intent and assistance 
      preference>",
      "example_bullet_ids": ["<existing bullet id>"]
    }
  ]
}
\end{verbatim}
~

$<$evolved\_memory$>$

\{memory\}

$<$/evolved\_memory$>$\\

Infer a unified user model. Return only the JSON object.
\end{promptbox}

\section{OSWorld-Persona Construction Details}
\label{appendix:dataset}

\begin{table}[t]
\centering
\resizebox{\textwidth}{!}{%
\begin{tabular}{llcccc}
\toprule
\textbf{Persona} & \textbf{Brief Description} & \textbf{Split} & \textbf{Examples} & \textbf{Support: Yes} & \textbf{Support: No} \\
\midrule
\multirow{2}{*}{Automation First}
& \multirow{2}{*}{\makecell[l]{Offloads repetitive/multi-step work to agents;\\hands-off for trivial one-off actions.}}
& Train & 521 & 202 & 319 \\
& & Test  & 239 & 100 & 139 \\
\midrule

\multirow{2}{*}{Eager Hands-on}
& \multirow{2}{*}{\makecell[l]{Delegates whole sub-tasks and reviews results;\\prefers doing over explaining.}}
& Train & 509 & 390 & 119 \\
& & Test  & 237 & 192 & 45 \\
\midrule

\multirow{2}{*}{Learning-oriented}
& \multirow{2}{*}{\makecell[l]{Prefers guidance and explanation over silent\\takeover;welcomes nudges when stuck.}}
& Train & 660 & 37 & 623 \\
& & Test  & 285 & 10 & 275 \\
\midrule

\multirow{2}{*}{Reserved Drafts}
& \multirow{2}{*}{\makecell[l]{Prefers AI-generated drafts or partial artifacts\\to review and edit independently.}}
& Train & 589 & 60 & 529 \\
& & Test  & 239 & 30 & 209 \\
\midrule

\multirow{2}{*}{\textbf{Total}}
& \multirow{2}{*}{/}
& Train & 2,279 & 689 & 1,590 \\
& & Test  & 1,000 & 332 & 668 \\
\bottomrule
\end{tabular}
}
\caption{\textbf{OSWorld-Persona statistics.} Annotations are split chronologically per persona into train and test sets to avoid leakage.
}
\label{table:data_statistics}
\end{table}

\begin{figure*}[t]
  \centering
  \includegraphics[width=0.95\textwidth]{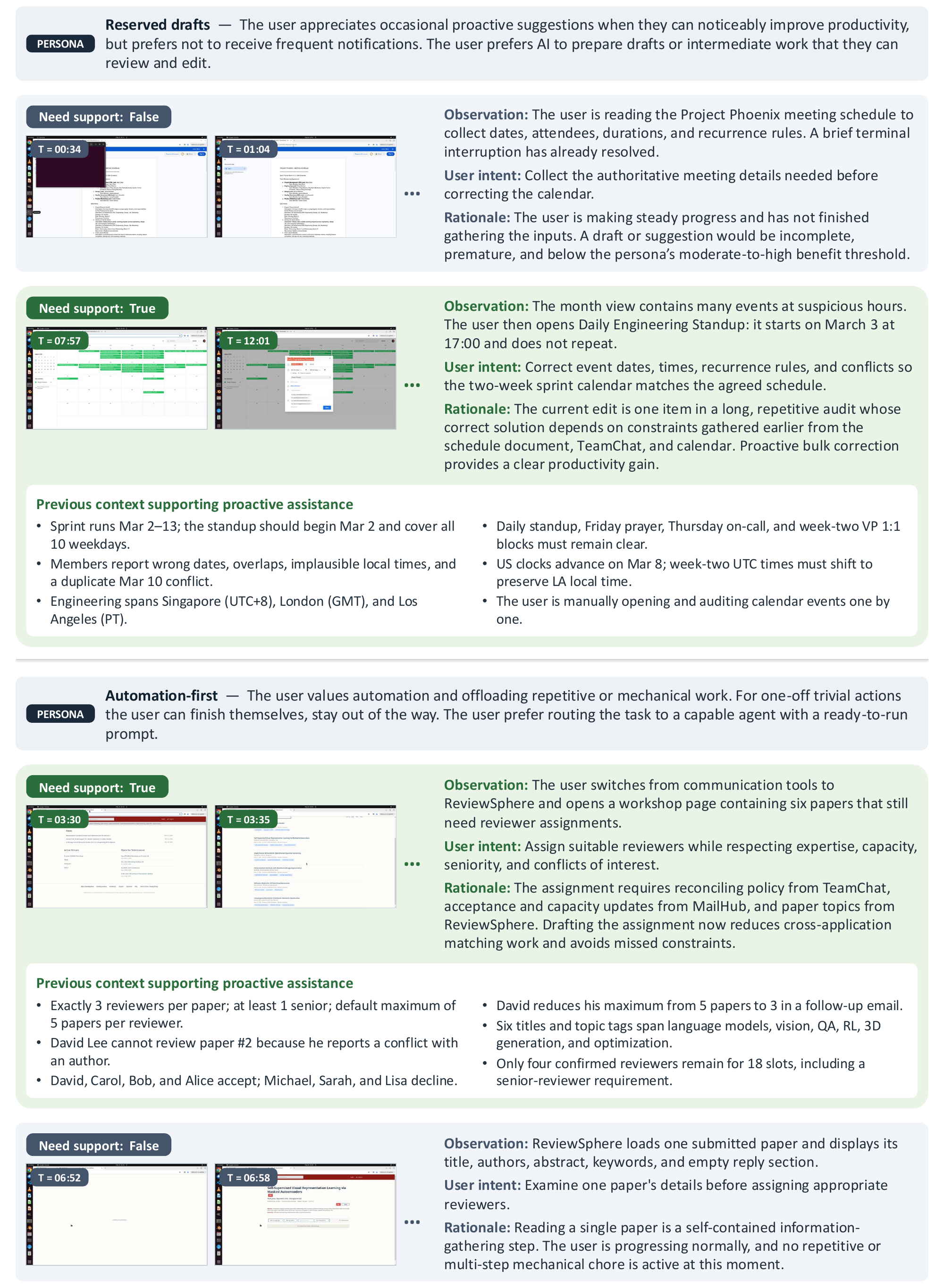}
\caption{\textbf{OSWorld-Persona data point examples.} Each data point contains screenshots from a computer-use session and persona-conditioned annotations of the observation, user intent, rationale, and \texttt{need\_support} label generated by Kimi-K2.6. After a self-consistency check, human annotators verify samples labeled \texttt{need\_support=True} and identify the preceding context that supports proactive assistance.
}
\label{fig:datapoint}
\end{figure*}

As discussed in~\refsec{sec:dataset}, we curate \textbf{OSWorld-Persona} through a five-stage data synthesis pipeline applied to computer-use sessions from OSWorld 2.0~\citep{yuan2026osworld2}. Here, we document the data synthesis details: \textbf{(1) Screen recording collection.} We use OSWorld 2.0 as the source computer-use dataset, which provides 108 long-horizon, complex computer tasks spanning both personal and professional scenarios. \textbf{(2) Key frame extraction.} From each session recording, we select representative key frames following the strategy used in efficient deployment that capture meaningful state transitions with threshold $\tau$ set to 2{,}000. \textbf{(3) Input instance construction.} We pair key frames in a rolling context window of up to five frames with a sampled user persona, which describes the user's preferences and behavioral patterns. \textbf{(4) Two-stage support-decision labeling.} The combined inputs are fed to Kimi-K2.6\footnote{\url{https://www.kimi.ai/ai-models/kimi-k2-6}} acting as the annotation VLM.  In the first stage, the VLM converts each window into a profile-agnostic observation and inferred user intent; in the second stage, it makes a persona-conditioned binary support decision. We show the prompts used for both labeling stages below.

To reduce stochastic label noise and ensure dataset quality, we apply a self-consistency check at temperature $0.4$ where each decision stage is repeated two more times with identical inputs and an example is retained only when both additional runs unanimously reproduce the original \texttt{need\_support} label. We further apply human validation by letting human to review cases where \texttt{need\_support} is true and annotating context that informs potential proactive support; the data point will be removed if the human annotator cannot find context supporting the proactive support. \textbf{(5) Chronological train/test split.} Labeled sessions are split chronologically per user with 70\% as train and 30\% as test, ensuring the test set faithfully reflects the natural temporal distribution of real computer-use workloads. \reftab{table:data_statistics} reports the detailed dataset statistics and \reffig{fig:datapoint} illustrates data point examples.

\begin{promptbox}{Two-stage Labeling Stage 1: Perceive}
\small
You are a screen-understanding module for a proactive desktop assistant.

You are shown the current Screen History: a handful of key frames in chronological order (deduplicated, so consecutive frames reflect real changes). A \texttt{screenshots} block lists each frame as "[t=mm:ss] Screenshot i of N" with its in-video timestamp; the images follow that text in the same order. Use the timestamps to reason about elapsed time, repetition, and whether the user is progressing or stalling.

You may also be given a Recent Observation: what was observed in the moments just before this window. Use it for continuity — to judge whether the user is making progress, repeating a failed approach, or has moved on — but your observation must describe the CURRENT Screen History, not restate the past.

Your ONLY job here is to describe what is happening on screen — not to decide whether to help. Produce a faithful, concrete observation grounded in what is visible across the frames (which application, what the user is doing, any errors, dialogs, repeated actions, or signs of being stuck or making progress), and infer the user's immediate intent.

Respond with ONLY a JSON object, no prose, no code fences:
\begin{verbatim}
{
  "observation": "<2--4 sentences describing what unfolds
    across the current frames, citing visible evidence and
    timestamps>",
  "user_intent": "<the user's immediate goal; use a string,
    or a JSON list of strings if there are clearly several>"
}
\end{verbatim}
\end{promptbox}

\begin{promptbox}{Two-stage Labeling Stage 2: Decide}
\small
You are the decision module of a personalized proactive desktop assistant.

You are given:
\begin{itemize}
    \item {a User Preference Memory describing when and how this user wants help,}
    \item {an Available Agent Profile listing the tools/agents you can route work to,}
    \item {an observation of what is currently happening on the user's screen,}
    \item {the user's inferred intent, and}
    \item {optionally a Recent Observation: what was happening just before now.}
\end{itemize}

Decide whether to proactively support the user RIGHT NOW, honoring the user's preferences. Be conservative: only set \texttt{need\_support}  to "yes" when the expected benefit clears the bar the user's preferences set, and stay silent ("no") when the user is making steady progress, the action is trivial, or an interruption would not be worth it. Use the Recent Observation to weigh continuity — e.g. a user who has been stuck or repeating the same failed action across observations is a stronger case for help than a momentary pause; do not re-offer help that the recent context shows would just repeat itself.

\texttt{suggestion\_type} must be one of:
\begin{itemize}
    \item{\texttt{none}: Provide no support. In this case,
    \texttt{suggestion} must be an empty string.}
    \item{\texttt{direct\_message}: Provide a short message directly to the user, such as concise advice, a ready-to-use command, or a draft snippet.}
    \item{\texttt{prompt\_to\_agent}: Provide a ready-to-run instruction for an agent to complete the task. Choose this option only when an appropriate agent is listed in the Available Agent Profile. Identify the agent and write the exact prompt that should be sent to it.}
\end{itemize}

Respond with ONLY a JSON object, no prose, no code fences:
\begin{verbatim}
{
  "need_support": "yes" | "no",
  "rationale": "<why this decision, explicitly referencing the 
  user's preferences and the observed situation>",
  "suggestion_type": "none" | "direct_message" | "prompt_to_agent",
  "suggestion": "<the message or agent prompt; empty string if
  need_support is no>"
}
\end{verbatim}
\end{promptbox}


\clearpage
\section{Profiling On-device Deployment}
\label{appendix:profiling}

\begin{figure}[t]
    \centering
    \includegraphics[width=\linewidth]{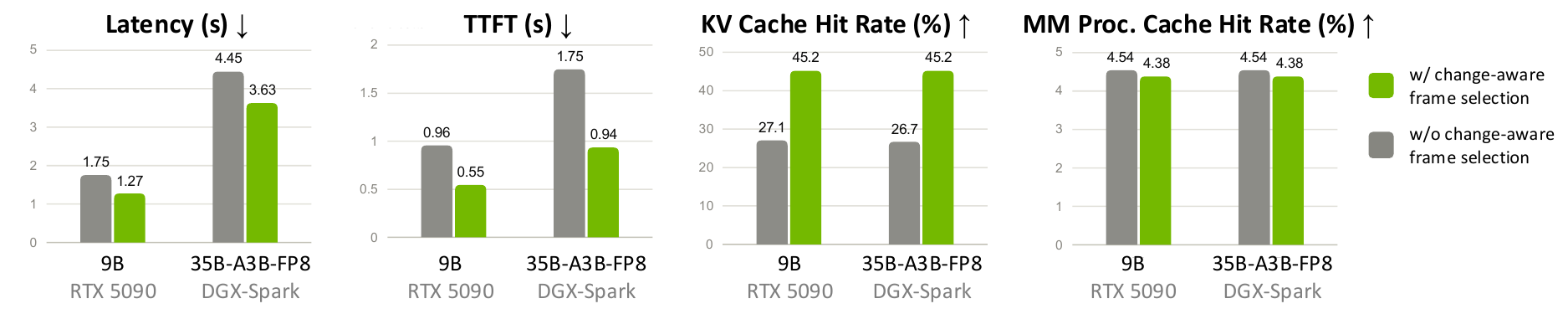}
    \caption{\textbf{Effect of change-aware frame selection on a replayed computer-use workload.} We profile two on-device deployment configurations: Qwen3.5-9B on an RTX 5090 and Qwen3.5-35B-A3B-FP8 on DGX Spark. For each configuration, we compare Observer Agent workloads with and without change-aware key-frame selection.}
    \label{fig:observer_input_strategy_comparison}
\end{figure}

\begin{table}[t]
\centering
\resizebox{\textwidth}{!}{%
\begin{tabular}{llcccccccccccc}
\toprule
\multirow{2}{*}{\textbf{Model}} & \multirow{2}{*}{\textbf{Device}}
& \multicolumn{3}{c}{\textbf{Latency (s)}}
& \multicolumn{3}{c}{\textbf{TTFT (s)}}
& \multicolumn{3}{c}{\textbf{KV Cache Hit \%}}
& \multicolumn{3}{c}{\textbf{MM Proc. Cache Hit \%}} \\
\cmidrule(lr){3-5}
\cmidrule(lr){6-8}
\cmidrule(lr){9-11}
\cmidrule(lr){12-14}
& & \textbf{Mean} & \textbf{P50} & \textbf{P95} & \textbf{Mean} & \textbf{P50} & \textbf{P95} & \textbf{Mean} & \textbf{P50} & \textbf{P95} & \textbf{Mean} & \textbf{P50} & \textbf{P95} \\
\midrule
Qwen-3.5-9B & RTX5090
& 2.84 & 2.42 & 6.97
& 1.22 & 0.94 & 3.68
& 29.23 & 39.74 & 55.97
& 17.57 & 20.00 & 50.00 \\
Qwen-3.5-35B-A3B-FP8 & DGX Spark
& 7.52 & 6.63 & 18.68
& 2.83 & 1.70 & 10.49
& 27.92 & 40.50 & 55.87
& 26.29 & 25.00 & 75.00 \\
\bottomrule
\end{tabular}
}
\caption{\textbf{Profiling results under a real single-user workload with change-aware frame selection enabled.} Requests include accumulated context and user preferences learned with Coco-PE.}
\label{table:real_workload_profiling}
\end{table}

To support Coco's core component, the Observer Agent, on local AI computing platforms such as DGX Spark, we used \texttt{CTI}\footnote{\url{https://github.com/mit-han-lab/cti-engine}}, a single-GPU engine that supports both inference and fine-tuning on the same model instance. Unlike vLLM~\citep{kwon2023efficient} and SGlang~\citep{zheng2024sglang}, which are primarily designed for inference, \texttt{CTI} integrates inference and training into a unified system. This design allows idle compute resources to be used for fine-tuning according to the agent's needs while avoiding the additional memory overhead of maintaining separate serving and training systems, making it better suited to memory-constrained platforms and the deep personalization workload in Coco.

Built on Mini-SGLang, \texttt{CTI} retains a lightweight codebase. On top of \texttt{CTI}, we introduce two optimizations tailored to Coco's workload and the hardware characteristics of DGX Spark. First, to mitigate the memory-bandwidth bottleneck of DGX Spark, we enable Qwen3.5 DFlash~\citep{chen2026dflash} speculative decoding with a draft block size of 8 to accelerate the decoding process. Second, because Coco's requests share highly repetitive system-prompt prefixes, we employ a radix prefix cache that matches prefixes at a 16-token page granularity and checkpoints both the Qwen3.5 recurrent state and the DFlash draft state at eligible shared-prefix boundaries. Together, these optimizations accelerate decoding, enable effective prefix reuse, and reduce redundant prefill computation.

With the aforementioned \texttt{CTI} setup, we profile Coco's on-device deployment using two hardware-model configurations: Qwen3.5-9B on an RTX 5090 and Qwen3.5-35B-A3B-FP8 on DGX Spark. The former represents a high-bandwidth consumer GPU configuration, while the latter leverages DGX Spark's larger memory capacity to run a substantially larger model despite its lower memory bandwidth. We evaluate these configurations along two dimensions: first, using a controlled replay workload to isolate the efficiency gains from change-aware frame selection; and second, using a real single-user workload to measure end-to-end Observer Agent performance with accumulated context and learned user preferences.

\noindent
\textbf{Replay Workload}\quad
We first evaluate different visual input strategies using a controlled replay workload. We record one 30 minutes of computer-use activity from one author and construct requests by aggregating user actions within 15-second windows. Windows containing fewer than two mouse or keyboard actions are merged with the subsequent window, resulting in 80 Observer Agent requests. This replay workload is designed specifically to profile computer-use inputs and therefore excludes accumulated context and learned user preferences. We compare change-aware frame selection, using an MSE threshold of $\tau=8,000$, against a baseline that naively includes all frames within each window. As shown in \reftab{table:real_workload_profiling}, change-aware frame selection substantially improves prefix reuse and inference latency. The KV cache hit rate increases by over $1.6\times$ under both deployment configurations: from 27.1\% to 45.2\% for Qwen3.5-9B on RTX 5090 and from 26.7\% to 45.2\% for Qwen3.5-35B-A3B-FP8 on DGX Spark. Correspondingly, Time-To-First-Token (TTFT) improves more than 1.7$\times$.

\textbf{Real Single-user Workload}\quad
We further profile on-device deployment under a real single-user workload, where each request includes accumulated context $h_t$ and user preference learned through Coco-PE. On the RTX 5090 configuration, we observe a mean TTFT of 1.22s, a mean end-to-end Observer Agent latency of 2.84s, and a KV cache hit rate of 29.23\%. On DGX-Spark, the corresponding values are 2.831s, 7.518s, and 27.92\%. \reftab{table:real_workload_profiling} reports P50 and P95 statistics.

These results show that the Observer Agent can operate at interactive timescales on local hardware. In particular, the observed latency is compatible with Coco's interaction cycle, in which the Observer Agent processes user activity over multi-second windows rather than responding to individual low-level actions. This demonstrates the feasibility of providing real-time, personalized assistance while keeping continuous computer-use data on device.

\section{Experiment Details}
\textbf{Coco-PE Implementation Details}\quad
For each persona, Coco-PE initializes an empty textual memory in the Observer Agent prompt. Examples are processed in batches of 8, with up to 8 screenshots each. At each batch, the current memory is used to generate predictions; all incorrect predictions and at most two adjacent correct predictions are passed to the reflector, and the curator applies at most 8 add, update, or delete operations. Reflection and curation use a temperature of 0.2, with output limits of 8,192 and 4,096 tokens, respectively. In the self-evolved Observer Agent prompt, the memory is capped at 50 bullets, with lower-utility bullets removed when the cap is exceeded. The prompt evolution epoch is set to 1.

\textbf{Coco-LoRA Implementation Details}\quad
We train a separate persona-specific LoRA adapter for each backbone using LLaMAFactory\footnote{\url{https://github.com/hiyouga/LlamaFactory}}. We use LoRA rank of 32, a scaling factor of $\alpha=64$, and zero LoRA dropout. We train the LoRA adapter on the training split for 1 epoch under BF16 precision with a batch size of 8, using a cosine learning-rate schedular with a 10\% warmup ratio. For Qwen3.5-9B and Qwen3.5-35B-A3B, we use a learning rate of $1e-5$; for Nemotron-3-Nano-Omni, we use a learning rate of $5e-5$; all with thinking disabled. The vision encoder and multimodal projector are frozen, while LoRA is applied to all eligible linear modules in the language model except the output head. For MoE backbones, adapters are additionally attached to expert and router/gating projections when those components are exposed as linear modules.

\section{Coco Desktop Interface}
\label{appendix:coco_interface}

\begin{figure}[t]
    \centering
    \includegraphics[width=\linewidth]{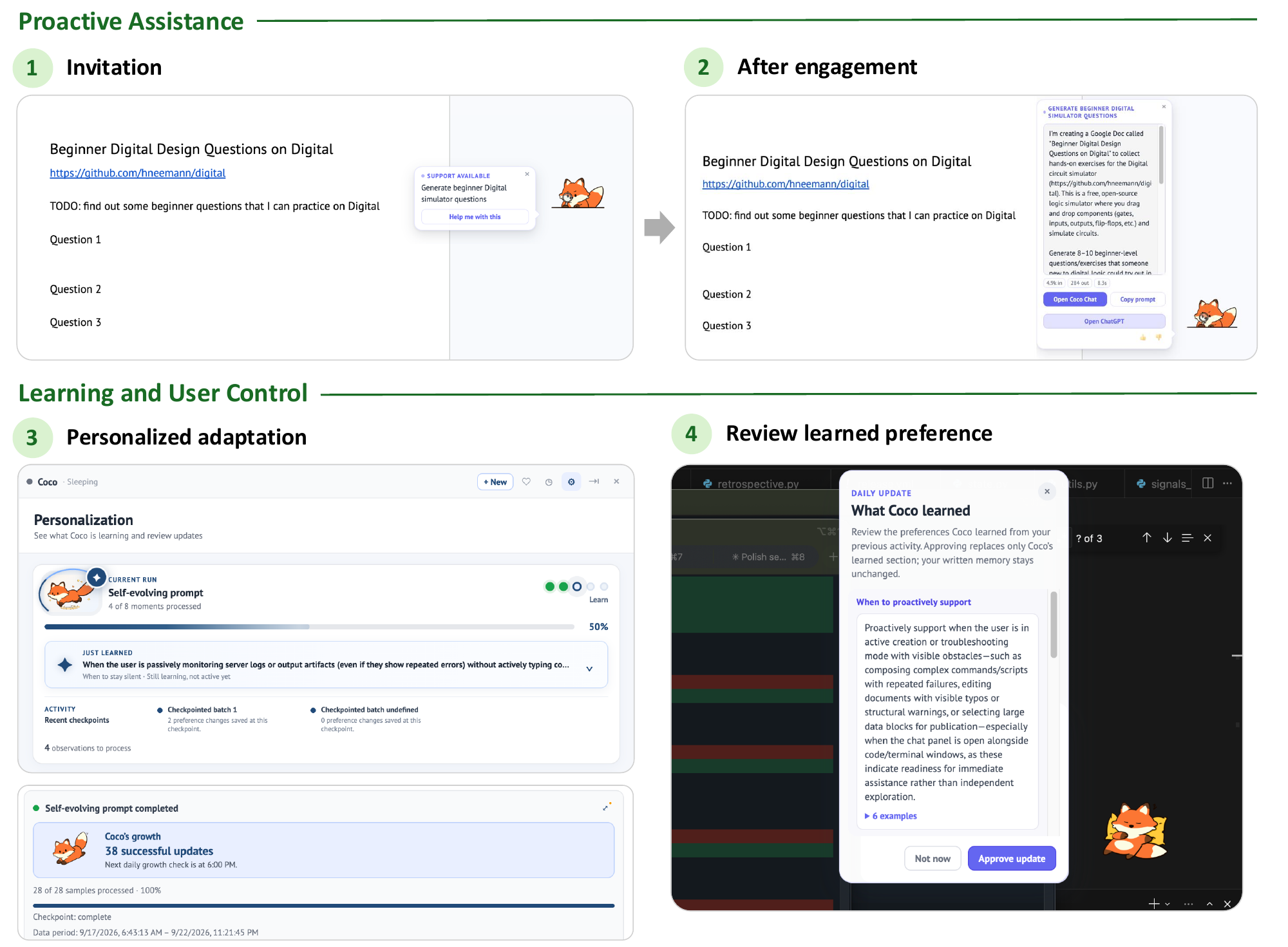}
    \caption{\textbf{Coco's desktop interface.} Top: Coco presents a brief invitation when it detects a proactive support opportunity and reveals the full suggestion only after the user engages. Bottom: the personalization view shows learning progress and lets users review learned preferences and their supporting examples.}
    \label{fig:coco_interface}
\end{figure}

Coco is implemented as a lightweight desktop companion that remains available without occupying a conventional application window. Its primary interface is a small, movable avatar displayed above the user's existing applications. The avatar provides an ambient indication that Coco is active while preserving the user's current workspace. Users may reposition or hide the avatar, review previous observations, or initiate a conversation through its action menu. \reffig{fig:coco_interface} illustrates Coco's desktop interface.

\textbf{Proactive Suggestions}\quad
Coco uses progressive disclosure to minimize interruption. When it identifies an opportunity to assist, it displays a compact pop-up next to the avatar containing only a short description of the opportunity and a ``Help me with this'' button. The complete suggestion is not shown unless the user explicitly clicks the button to engage. The initial pop-up can be dismissed immediately and disappears automatically if the user does not interact with it. When the avatar is hidden, the same interaction is presented as a small notification in the corner of the screen. If the user does not click ``Help me with this'' button, Coco views it as a negative reward to the corresponding suggestion. During our user study, this also counts as a false positive case. After the user engages, Coco reveals the full suggestion in place. Depending on the situation, this may be directly usable content returned from the Execution Agent, such as a summary or recommendation, or a delegation prompt containing the context needed for multi-turn agentic execution. Users can copy the result, continue discussing it in Coco's chat interface, or open one of their configured AI tools with the generated prompt. Users can provide lightweight positive or negative feedback using the thumbs-up and thumbs-down controls.

\textbf{User-initiated Assistance}\quad
Users do not need to wait for a proactive suggestion. Clicking the avatar opens a full conversation panel in which they can ask questions, provide follow-up instructions, and inspect Coco's responses. Coco can also be invoked with a keyboard shortcut, or through an optional wake-word interface. These entry points support text, voice, and screen-grounded requests while preserving the same conversational history. If a user requests assistance when the Observer Agent predicted no need for support, Coco treats that request as a negative reward for the prediction; in the user study, we count it as a false negative.

\paragraph{Activity and User Control}
An activity view summarizes Coco's recent observations and past suggestions, allowing users to revisit offered support and see when Coco identified a potential need for assistance. A personalization view shows the current learning stage, processing progress, recent checkpoints, and number of feedback signals incorporated. Users can inspect each proposed preference and the examples supporting it, then keep, edit, or reject it. A revised preference takes effect only after user review and approval, and the interface records when the latest update was applied. During the user study, preference editing was disabled so that the evaluation reflected the preferences produced by self-evolving personalization without user edits.

\end{document}